\documentclass[journal]{IEEEtran} 
\usepackage{url}
\usepackage{amsmath}
\usepackage{siunitx}
\usepackage{amssymb}
\usepackage{hyperref}
\usepackage{float}
\usepackage{balance}

\usepackage[numbers]{natbib}
\usepackage{listings}
\usepackage{algorithm}
\usepackage{algpseudocode}

\usepackage{makecell}
\usepackage{tabulary}
\usepackage{rotating}
\usepackage{booktabs}

\usepackage[export]{adjustbox}
\usepackage{graphicx}
\usepackage{xspace}
\usepackage{subcaption}
\usepackage{wrapfig}
\usepackage[normalem]{ulem}

\usepackage{tikz}
\usetikzlibrary{patterns}
\usetikzlibrary{calc}

\algnewcommand{\algorithmicparams}{\textbf{Parameters:}}
\algnewcommand\Params{\item[\algorithmicparams]}

\newcommand{\X}{X}
\newcommand{\Xcur}{X_{cur}}
\newcommand{\fiber}{F}

\newcommand{\R}{\mathbb{R}}
\def\Xk{\X_{k}}
\def\Xf{\X_\textnormal{free}}

\def\Xkk{\X_{k-1}}
\def\fiberk{\fiber_{k}}

\def\ManhattanAbbrv{MH\xspace}
\def\PriorityQueue{\ensuremath{\mathbf{X}}}

\def\lift{\textsc{Lift}}

\def\h{h}
\def\hstar{h^{*}}
\def\head{\ensuremath{H}}

\def\x{x}
\def\xi{\x_I}

\def\xg{\x_G}

\def\G{\ensuremath{\mathbf{G}}}

\def\path{\ensuremath{\mathbf{p}}}

\algnewcommand\True{\textbf{true}\xspace}
\algnewcommand\False{\textbf{false}\xspace}
\makeatletter
\newcommand{\toprulealg}{\hrule height.8pt depth0pt \kern2pt} 
\newcommand{\midrulealg}{\kern2pt\hrule\kern2pt} 
\newcommand{\bottomrulealg}{\kern2pt\hrule\relax}
\newcommand{\algcaption}[2][]{%
  \refstepcounter{algorithm}%
  \textbf{{\raggedright\fname@algorithm~\thealgorithm}}\ #2\par 
  \midrulealg
}
\makeatother

\def\basePath{p}

\def\xm{x_{\text{mid}}}

\def\pifk{\pi_{\fiber_k}}
\def\pif{\pi_{\fiber}}

\def\maxSample{\ensuremath{S_{\text{max}}}}
\def\maxDepth{\ensuremath{D_{\text{max}}}}
\def\maxBranch{\ensuremath{B_{\text{max}}}}
\def\deltaBase{\ensuremath{\delta_{\Xkk}}}

\def\deltaFiber{\ensuremath{\delta_{\fiberk}}}

\def\location{\ensuremath{\text{location}}}
\def\depth{\ensuremath{\text{depth}}}
\def\restriction{\ensuremath{\text{r}}}

\algnewcommand{\algorithmicbreak}{\textbf{break}}
\algnewcommand{\BREAK}{\algorithmicbreak}
\algnewcommand{\algorithmicor}{\textbf{ or }}
\algnewcommand{\OR}{\algorithmicor}
\algnewcommand{\algorithmicand}{\textbf{ and }}
\algnewcommand{\AND}{\algorithmicand}

\def\wrigglePattern{\textsc{WrigglePattern}\xspace}
\def\tunnelPattern{\textsc{TunnelPattern}\xspace}
\def\triplestepPattern{\textsc{TripleStepPattern}\xspace}
\def\L1Pattern{\textsc{ManhattanPattern}\xspace}

\def\xend{\x_\text{End}}
\def\xhead{\x_{\head}}
\def\xfiberhead{\x_{\fiber_{\head}}}
\def\lend{l_\text{End}}

\title{\Huge Section Patterns: Efficiently Solving Narrow Passage Problems in Multilevel Motion Planning}

\begin{document}

\author{Andreas Orthey$^{1}$ and Marc Toussaint$^{1,2}$
\thanks{$^{1}$Max Planck Institute for Intelligent Systems, Stuttgart, Germany. Marc Toussaint thanks the MPI-IS for the Max Planck Fellowship. {\tt\small {aorthey}@is.mpg.de}}
\thanks{$^{2}$Technical University of Berlin, Germany}
\thanks{\copyright 2020 IEEE.  Personal use of this material is permitted.  Permission from IEEE must be obtained for all other uses, in any current or future media, including reprinting/republishing this material for advertising or promotional purposes, creating new collective works, for resale or redistribution to servers or lists, or reuse of any copyrighted component of this work in other works.}
}

\maketitle

\begin{abstract}

Sampling-based planning methods often become inefficient due to narrow passages. Narrow passages induce a higher runtime, because the chance to
  sample them becomes vanishingly small. In recent work, we showed that narrow
  passages can be approached by relaxing the problem using admissible lower-dimensional
  projections of the state space. Those relaxations often increase the volume of narrow passages under
  projection. Solving the relaxed problem is often efficient and produces an
  admissible heuristic we can exploit. However, given a base path, i.e. a solution to a relaxed problem, there
  are currently no tailored methods to efficiently exploit the base path. To efficiently exploit the base path and thereby its admissible heuristic, we develop section patterns, which are solution strategies to efficiently
  exploit base paths in particular around narrow passages. To coordinate section patterns, we develop the pattern dance algorithm, which efficiently coordinates section patterns to reactively traverse narrow passages. We combine the pattern dance algorithm with previously developed multilevel planning algorithms and benchmark them on
  challenging planning problems like the Bugtrap, the double L-shape, an
  egress problem and on four pregrasp scenarios for a 37 degrees of freedom shadow hand mounted on a
  KUKA LWR robot. Our results confirm that
  section patterns are useful to efficiently solve high-dimensional
  narrow passage motion planning problems. 

\end{abstract}

\section{Introduction}

\begin{figure}
    \centering
    \includegraphics[width=0.95\linewidth]{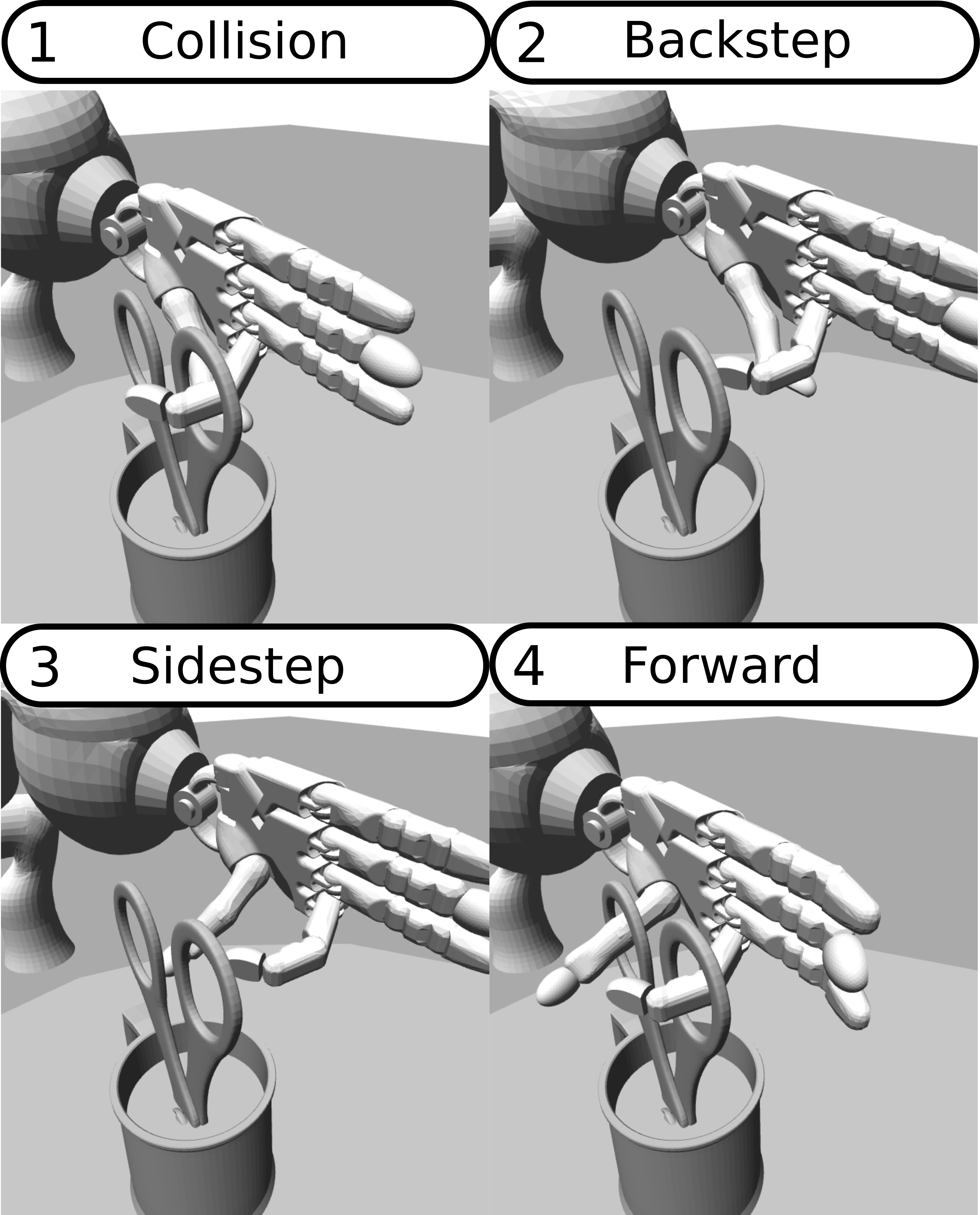}
    \caption{Efficient exploitation of admissible heuristics (stemming from solution to relaxed problem) using the triple step pattern. The triple step pattern is one of four section patterns we advocate to efficiently exploit admissible heuristics near narrow passages.}
    \label{fig:pullfigure}
\end{figure}

Sampling-based motion planning algorithms are a successful paradigm to automate robotic
tasks \cite{lavalle_2006}. However, sampling-based algorithms do not perform well when the state space of the robot contains narrow passages \cite{Mainprice2020, Szkandera2020, Hsu2003, Saha2005}, which are low-measure regions which have to be
traversed to reach a goal. Narrow passages are often occurring in tasks which are
particularly important in robotic applications, like grasping, peg-in-hole, egress/ingress or long-horizon planning problems \cite{Fu2019, Hartmann2020}.

In previous work, we and other research teams have shown that we can often efficiently solve high-dimensional
planning problems by using admissible
lower-dimensional projections of the state space, a topic we refer to as
multilevel motion planning \cite{Ferbach1997, Bayazit2005, Orthey2020IJRR, Reid2020, Vidal2019}. When
using a multilevel motion planning framework, we can often use solutions to
simplified planning stages as admissible heuristics for the original problem
\cite{Pearl1984, Aine2016}. To efficiently exploit those admissible heuristics, we can use biased sampling methods \cite{Orthey2020IJRR, Reid2019}, which we can combine with classical planning algorithms like the rapidly-exploring random tree algorithm
\cite{Orthey2019}, the probabilistic roadmap planner \cite{Orthey2018}, its
optimal star versions \cite{Orthey2020IJRR} or the fast marching tree planner
\cite{Reid2019}. However, while showing promising runtimes, those algorithms are prone to get trapped when run on problems involving narrow passages.

In this work, we address narrow passages in multilevel motion planning problems
by developing section patterns. Section patterns are methods to explicitly address problematic situations that occur when we exploit solutions to relaxed problems. We introduce four section patterns. First, we introduce the Manhattan pattern, which we use to compute solution paths which actuate the minimal amount of joints to reach a goal region, which is advantageous for high dimensional systems \cite{Cortes2008, Orthey2020IJRR}. 
Second, we introduce the Wriggle pattern, which we use to make small random walk steps to traverse a narrow passage. Third, we introduce the Tunnel pattern, which we use to steer around small infeasible regions. Fourth, we introduce the Triple step pattern, which we use to backtrack in case the algorithm gets stuck. 
In Fig.~\ref{fig:pullfigure} the Triple step pattern is showcased for a $37$-degree of freedom (dof) robotic hand. We execute the pattern when a collision occurs (1). We first backstep (2), then sidestep (3) and finally we make a forward step (4) to reach a goal position. The details of this and the other patterns will be detailed later in this paper.

To coordinate the execution of the four section patterns, we develop a novel algorithm we call \emph{pattern dance}. The pattern dance algorithm applies the section patterns sequentially by trying first a pattern which is easy to compute (Manhattan pattern) and reverting to the more complex pattern like Wriggle or Tunnel only if needed. If all those patterns fail, we revert to the Triple step pattern, which is the most computationally demanding pattern. We embed this pattern dance algorithm into four multilevel planners \cite{Orthey2020IJRR}, namely the quotient space RRT (QRRT) \cite{Orthey2019}, the quotient space roadmap planner (QMP) \cite{Orthey2018} and its optimal versions QRRT* and QMP* \cite{Orthey2020IJRR}.

Our contributions are as follows.
\begin{enumerate}
    \item We develop section patterns to efficiently exploit base space
      paths (solutions to relaxed problems).
    \item To coordinate sections patterns, we develop the pattern dance algorithm.
    \item We combine the pattern dance algorithm with four multilevel planners (QRRT, QRRT*, QMP, QMP*) and compare against $36$ planners from the open motion planning library (OMPL) and a previous sidestepping
      algorithm \cite{Orthey2020IJRR} on $7$ challenging scenarios.
\end{enumerate}

\section{Related Work}

Let us review the literature by focusing on two topics. First, we focus on generating admissible heuristics \cite{Edelkamp2011} for motion planning problems involving continuous domains \cite{lavalle_2006}. We discuss sources of admissible heuristics like constraint relaxations, lazy search, informed trees and past experience. Second, given an admissible heuristic, we review methods to efficiently exploit the heuristic either using path section approaches, local minima avoidance or narrow passage handling. 

\subsection{Generating Admissible Heuristics}

Motion planning \cite{lavalle_2006} is a well studied topic which has been successfully applied to a wide range of problem domains \cite{Moll2015}. One of the most promising paradigms to solve motion planning problems are (asymptotically optimal) sampling-based planners \cite{Karaman2011, Salzman2016, Salzman2019, Bekris2020, Gammell2020Survey}.  However, these planners might become inefficient in state spaces which are too high-dimensional \cite{Orthey2020IJRR}, contain intricate constraints \cite{Jaillet2012} or narrow passages \cite{Lee2012}. We can, however, often solve such problems efficiently, if we use admissible heuristics \cite{Aine2016}. 

We believe there are three large sources of admissible heuristics. First, we can compute admissible heuristics as solutions to relaxed problems \cite{Pearl1984}. Early instances of this idea to motion planning can be found in the constraint relaxation frameworks by \citeauthor{Ferbach1997} \cite{Ferbach1997}, \citeauthor{Sekhavat1998} \cite{Sekhavat1998} and \citeauthor{Bayazit2005} \cite{Bayazit2005}. Newer instances of this idea are putting the focus on different aspects like the specific type of projection \cite{Sucan2011, Gochev2012} or the type of lower-dimensional space \cite{Orthey2018, Brandao2020}. We refer to all those frameworks under the collective term multilevel motion planning \cite{Orthey2020IJRR}. We can apply multilevel frameworks both to holonomic \cite{Reid2019, Reid2020} and nonholonomic planning problems \cite{Vidal2019, Orthey2020IJRR}. To create multilevel abstraction, we can often remove links from a robot \cite{Bayazit2005, Zhang2009}, shrink links \cite{Baginski1996, Saha2005} or approximate a robot by simpler geometries, either exact \cite{Orthey2018, Grey2017} or approximate \cite{Brock2001, Rickert2014, tonneau_2018}.
While most methods use prespecified levels of abstraction, we can also use workspace information to compute abstractions on the fly \cite{Yoshida2005, Luna2020}, adaptively switch between abstractions \cite{Styler2017} or learn useful abstractions for specific instances \cite{Brandao2020}. Our approach is similar, in that we also use a multilevel motion planning framework \cite{Orthey2020IJRR}. However, our work is complementary, in that we focus specifically on computing path sections in the presence of narrow passages in the state space.

A second source of admissible heuristics are lazy search \cite{Bohlin2000, Haghtalab2017} and informed sets \cite{Gammell2014, Joshi2020}. Instead of using relaxations, we can compute lazy paths (paths not checked for collisions), either forward from the start \cite{Hauser2014} or backwards from the goal \cite{Strub2020Adaptively}, to create an efficient heuristic which we can exploit using dedicated algorithms \cite{Gammell2020}. Once a solution exists, we can also exploit informed sets, sets which exclude all states with provable higher cost-to-go \cite{Gammell2014, Gammell2020}. Those methods are particularly important, since edge evaluations is one of the bottlenecks in motion planning \cite{Kleinbort2020}. It therefore makes sense to develop heuristics which evaluate edges as late as possible \cite{Mandalika2019, Hou2020}.

Third, inspired by pattern database approaches in discrete search \cite{Culberson1998, Edelkamp2012, Hu2019}, we can also construct admissible heuristics by using past experience. We can achieve this by either precomputing motion primitives, like steering functions or controllers like linear quadratic regulators \cite{Sakcak2019Auto, Sakcak2019}. Or, we can store previous solution paths directly and use them as heuristics in new environments \cite{Driess2020, Qureshi2020}. Our work is complementary in that we assume a heuristic given and we focus on exploiting this heuristic as efficiently as possible.


\subsection{Exploiting Admissible Heuristic}

Given an admissible heuristic, we can optimally exploit it by discretizing the state space \cite{Ferguson2005} and by using the A* algorithm \cite{Hart1968, Pearl1984, Aine2016}. However, discretizing the state space usually does not scale well to higher dimensional state spaces \cite{Bungartz2004, Persson2014, Giles2015} and performance would be sensitive to the resolution used \cite{Du2020}. To avoid discretization, we found three categories of work which use continuous methods to exploit admissible heuristics.

First, we can use biased sampling methods. A straightforward way would be to represent the heuristic value of a state by the radius of a hypersphere around the state \cite{Littlefield2018}. We could then exploit this hypersphere using dynamic domain sampling \cite{Yershova2009}. Using such a scheme, we would expand states with higher heuristic values more often. Depending on the exact type of heuristic function used, we would obtain sampling distributions which would increase the probability to sample states which are near to restricted workspace geometries \cite{VanDenBerg2005, Yang2005}, to state space obstacles \cite{Amato1998} or to narrow passage \cite{Hsu2003}. Those sampling distributions could also be learned over time to improve sampling \cite{Luo2019, Ichter2018}. Our approach is similar in that we also use sampling-based methods. We differ, however, in that we concentrate on designing efficient patterns complementary to biased sampling methods.

Given a solution to a relaxed problem, we can often use this solution as a guide path heuristic \cite{Zhang2009, tonneau_2018} to quickly find a solution in the original state space. Using the parlance of fiber bundles, we call this the find section problem \cite{Orthey2020IJRR}. This problem requires a relaxed solution (a base path), which we can find by computing workspace regions \cite{Plaku2007}, by using workspace graphs \cite{Denny2020, Uwacu2020} or by using a simpler robot geometry \cite{tonneau_2018}. In more complex environments, it is often advantageous to use multiple base paths \cite{Vonasek2019, Denny2020} which decompose the original problem into smaller subproblems \cite{pokorny_2016_ijrr, bhattacharya_2018, Orthey2020WAFR}. To exploit a base path, we can often use restriction sampling \cite{Palmieri2016, Orthey2018}, which is highly efficient in high-dimensional state spaces, where uniform sampling would most likely fail to find solutions in a reasonable time \cite{Grey2017}. Apart from biasing sampling, we can also explicitly search over the set of states which project onto the base path \cite{Zhang2009}, which we call the path restriction. To find paths over path restrictions, we previously developed a sidestepping approach \cite{Orthey2020IJRR}, where we propagate states along the path restriction and execute sidesteps when collision occur. However, as we show in Sec.~\ref{sec:sectionpatterns}, sidesteps are often not beneficial for narrow passages. While we also search over path restrictions, we differ by developing dedicated patterns to more efficiently traverse narrow passages.

Path section approaches and other heuristic search methods often fail because they reach local minima. We define a local minimum as a region in state space where the heuristic is not or only weakly correlated with the true cost-to-go \cite{Vats2017}. To address local minima, we can choose one of two approaches. First, we could preemptively avoid local minima. If the environment is static, we can learn minima regions and use this information to update the heuristic function \cite{Vats2017}. Second, we could try to escape local minima. There exist several methods to escape local minima like deflating the heuristic value of states close to obstacles \cite{Du2019} or increasing the search resolution to prevent evaluation of closeby states \cite{Du2020}. A related idea is to utilize Tabu search \cite{Glover1998} to prevent sampling
in previously visited regions. 

It is important to make the distinction between local minima which trap the planner and regions which might look like local minima but which a planner can actually traverse. We call such regions narrow passages \cite{Salzman2013}. To verify the existence of narrow passages in low-dimensional state spaces, we can use exact infeasibility proofs \cite{Schweikard1998, Basch2001}, for example using geometrical shapes like alpha complices \cite{Mccarthy2012} or cell decomposition methods \cite{Zhang2008}. Because many state spaces have a local product structure, we can often use configuration space slices \cite{Lozano1987, Sintov2020} to efficiently test for infeasibility \cite{Varava2020}. If the problem is feasible, we could then use the geometrical shapes to enumerate narrow passages \cite{Manak2019}. To exploit narrow passages, we could bias sampling to the most constricted areas \cite{Yang2005, Szkandera2020}. We differ to those approaches by not explicitly modeling narrow passages or local minima, but we instead develop reactive measures to escape minima and to traverse narrow passages. We thereby avoid spending time on irrelevant narrow passages.

\section{Background\label{sec:background}}

\begin{figure}
    \centering
    \includegraphics[width=\linewidth]{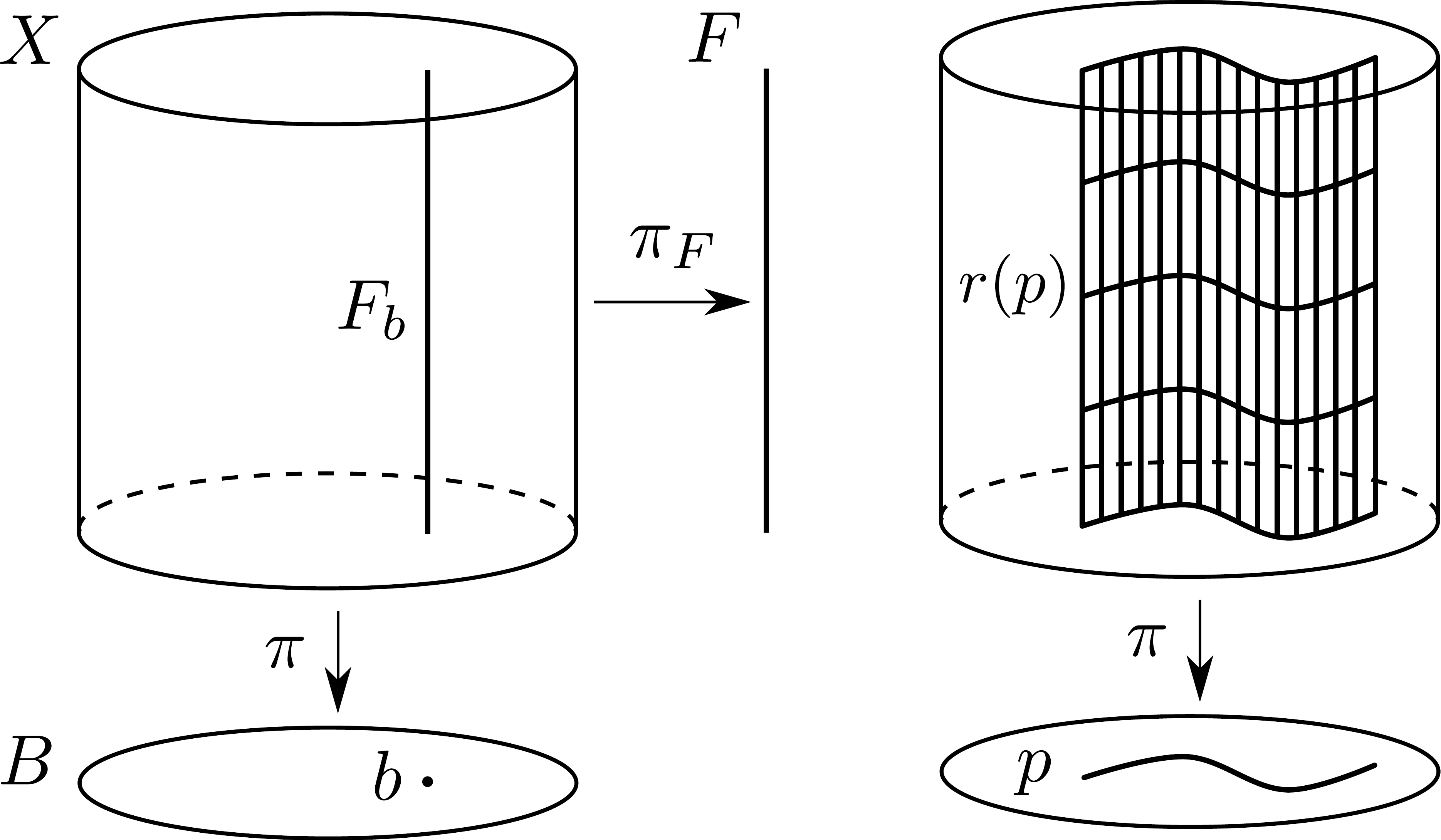}
    \caption{
    Left: Fiber bundle $\R^3 \rightarrow \R^2$ with base space $B$, total space $X$, fiber space $F$, mappings $\pi$, $\pi_F$ and fiber $F_b$ over base element $b$. Right: Path restriction $r(p)$ over base path $p$. Adapted from \cite{Orthey2020IJRR}.\label{fig:fiberbundle}}
\end{figure}

Let us describe the necessary background to follow the exposition of our algorithm in Sec.~\ref{sec:algorithm} and Sec.~\ref{sec:sectionpatterns}. We start by explaining multilevel motion planning, i.e.~planning with sequences of relaxed subproblems. While several formulations exist, we believe the framework of fiber bundles \cite{Orthey2020IJRR} to be a good way to concisely model multilevel abstractions and describe our algorithms. We then describe the concepts of lift, path restriction and path section which are particularly important. 
Finally, we describe the notion of admissible heuristics, which is one of the fundamental concepts to exploit solutions to relaxed problems \cite{Pearl1984}.

\subsection{Optimal Motion Planning}

Let $\X$ be the state space of the robot. To each state space we associate a constraint function $\phi: \X \rightarrow \{0,1\}$ which evaluates to $0$ if a state is constraint-free and to $1$ otherwise. We use the constraint function to define the free state space $\Xf = \{ x \in \X \mid \phi(x) = 0\}$. Together with an initial configuration $\xi \in \Xf$ and a goal configuration $\xg \in \Xf$, we define an optimal motion planning problem \cite{Karaman2011, Salzman2019, Bekris2020} as the tuple $(\Xf, \xi, \xg, c)$, whereby our task is to develop an algorithm which computes a path from $\xi$ to $\xg$ while staying in $\Xf$ and minimizing the cost functional $c$. In this work, we use a minimal-length cost functional, but other costs are also possible like minimal energy or maximum clearance.


\subsection{Multilevel Motion Planning\label{sec:background:multilevel}}

Since high-dimensional motion planning problems are often too computationally expensive to solve, we use a sequence of relaxed problems which we refer to as multilevel abstractions
\cite{Orthey2020IJRR}. Given a state space $\X$, let us denote a multilevel abstraction as the tuple $\{\X_1, \cdots, \X_K\}$ with $\X_K = \X$. To each state space $\X_k$, we associate a constraint function $\phi_k$ and a projection $\pi_k$ from $\X_k$ to $\X_{k-1}$. We say that the projection $\pi_k$ is admissible (w.r.t. the constraint functions), if $\phi_{k-1}(\pi_k(x)) \leq \phi_k(x)$ for any $x$ in $\X_k$. With admissibility, we basically guarantee that solutions are preserved under projections \cite{Orthey2019}. If we would allow inadmissible projections, we would potentially sacrifice solutions and thereby sacrifice (probabilistic) completeness. 

\subsection{Fiber Bundle Formulation}

When working with multilevel abstraction, we quickly stumble upon situations where we lack the appropriate vocabulary to describe solution strategies. As a remedy, we describe multilevel abstractions using the framework of fiber bundles \cite{steenrod_1951, husemoller_1966, lee_2003}. A fiber bundle is a tuple $(\X_k, \X_{k-1}, \fiber_k, \pi_k, \pifk)$ consisting of a total space $\X_k$, a base space $\X_{k-1}$, a fiber space $\fiber_k$, a projection mapping $\pi_k$ from total to base space and a fiber projection mapping $\pifk$ from total to fiber space. We assume the projection mapping $\pi_k$ to be admissible. With a fiber bundle, we model product spaces which locally decompose as $\X_k = \X_{k-1} \times \fiber_k$. The total space $\X_k$ is a union of fiber spaces which are parameterized by the base space $\X_{k-1}$. If the level $k$ is unimportant for the task as hand, we often refer to a fiber bundle as the tuple $(\X, B, \fiber, \pi, \pif)$ with $\X$ being the total, $B$ the base, $\fiber$ the fiber space and $\pi$, $\pif$ the base and fiber projection, respectively. We visualize a prototypical fiber bundle in Fig.~\ref{fig:fiberbundle} (left). For more details and motivation, we refer to our prior work \cite{Orthey2020IJRR}. For the purpose of this paper, we focus on the three concepts of lift, path restriction and path section, which we explain next.

\subsection{Lift}

Let $(\X, B, \fiber, \pi, \pif)$ be a fiber bundle and let $b \in B$ be a base space element. We often like to project the element $b$ back to the total space $\X$. We call this operation a lift \cite{Roewekaemper2013, Orthey2020IJRR}. We define a lift as a mapping $\lift: B \rightarrow \X$. To uniquely select an element in $\X$, we will overload this function as a mapping $\lift: B \times F \rightarrow \X$ by providing a fiber space element $f$ in $\fiber$. If $\X$ is a product space, we define the lift as $\lift(b,f)=(b,f)$ \cite{Orthey2020IJRR}.

\subsection{Path Restrictions}

Let $p: I \rightarrow B$ with $I = [0,1]$ be a path on the base space (a base path). Given a base path, one of the most central sets which we use in this work are path restrictions. A path restriction is the set $r(p) = \{ x \in \X \mid \pi(x) \in p[I]\}$, whereby $p[I] = \{ p(t): t \in I \}$ is the image of the base path in $B$ and $\pi$ is the projection from $\X$ to $B$. We visualize this situation in Fig.~\ref{fig:fiberbundle} (right), where we show the image of a base path on the disk-shaped base space and its associated path restriction on the total space.

\subsection{Path Sections}

Given a path restriction, we are often interested in finding paths which are lying inside the path restriction. We call them path sections \cite{steenrod_1951}. A (smooth) path section w.r.t. a base path $p$ is a continuous mapping $s$ from base space $B$ to total space $\X$ such that $\pi(s(u))=u$ for any $u$ in the image of $p$ \cite{lee_2003}. This means, for each base path element, we select a unique state from the path restriction---in a continuous manner. 

\subsection{Admissible Heuristics}

Our motivation to introduce path restrictions and path sections comes from the role they play in exploiting admissible heuristics. Given a goal state $\xg$, an admissible heuristic $\h(x)$ for a state $x$ in $\X$ is a lower-bound on the true cost-to-go (or value) function $\hstar(x)$, which we define as the cost of the optimal path from $x$ to $x_G$ through $\Xf$. Formally, we write this condition as $h(x) \leq \hstar(x)$  \cite{Pearl1984, Aine2016, Orthey2019}. 

Given an admissible heuristic, we can try to reach the goal $\xg$ by using locally optimal decisions \cite{Hart1968}. If we are at a state $x$, we can make an optimal decision by doing a two-step approach. First, we compute the $f$-value of all its neighbors, which is the sum of its heuristic value and its cost-to-come from the start state. We then expand the state (node) with the lowest $f$-value, because, under the admissible heuristic, it is our best guess to efficiently reach the goal \cite{Pearl1984}.

However, in a continuous domain, we cannot straightforwardly compute all neighboring states. Instead, we imagine computing a small $\epsilon$-neighborhood around the state. To compute heuristic values, we project the complete neighborhood down onto the base space. To reach the goal, our best guess is to make a step into the direction of the current minimal-cost base path. The states which we would expand in that way are exactly the states on the path restriction. By searching a path section over this path restriction, we efficiently exploit the admissible heuristic given by the base path.
\section{Find Sections using Pattern Dance\label{sec:algorithm}}

Our goal is to develop an algorithm which solves the find section problem, the problem of finding a path section over a given path restriction. After we state the problem, we discuss how the problem fits into the more general framework of motion planning using multilevel abstractions \cite{Orthey2020IJRR}. Finally, we discuss the pattern dance algorithm, which coordinates four section patterns to efficiently find feasible path sections.


\subsection{Find Section Problem}

Let $(\X, B, \fiber, \pi, \pif)$ be a fiber bundle on $\X$ (possibly in a sequence of fiber bundles) and let $p: I \to B$ be a base path on $B$ starting at $\pi(\xi)$ and ending at $\pi(\xg)$. Given the base path $p$ and its path restriction $r(p) \subseteq \X$, our goal is to develop an algorithm to find a feasible path section, i.e. a path lying in the intersection of the path restriction $r(p)$ and the free state space $\Xf$ connecting $\xi$ to $\xg$. We call this problem the \emph{find section problem}. 

\begin{figure}
    \centering
    \includegraphics[width=\linewidth]{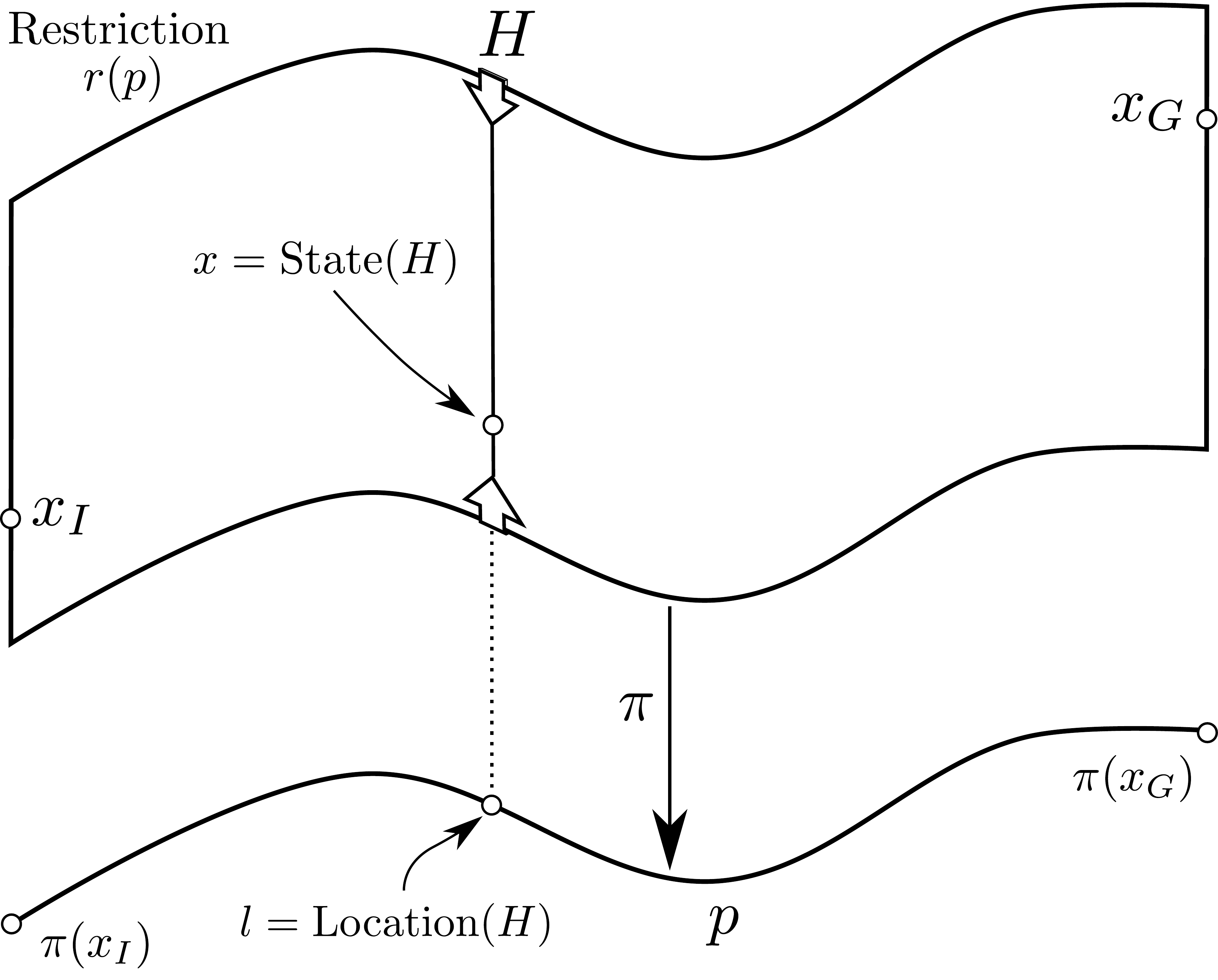}
    \caption{Path restriction $r(p)$ on a total space $\X$ over a base path $p$ from base space $B$, together with initial state $x_I$, goal state $x_G$, projection $\pi$ and head pointer with head pointer $\head$, consisting of state $x$ and location $l$.\label{fig:pathrestriction}}
\end{figure}

To illustrate the find section problem, we visualize it in Fig.~\ref{fig:pathrestriction}. The figure shows a base path $p$ on $B$ (bottom) and its restriction $r(p)$ on $\X$ (top). Our goal is to connect $\xi$ to $\xg$ while staying inside $r(p)$. To efficiently solve the find section problem, we often need to track information along the path restriction. To track this information, we introduce the notion of a head pointer $\head$ as the tuple $\head=(x, l, r)$ consisting of a path restriction $r(p) \subseteq \X$ over a base path $p$ in $B$, a current state $x$ in $r(p)$ and a location $l \in [0,1]$ defining the position along the base path. We think of the head pointer as a ruler which we move forward along the path restriction towards the goal state. In pseudocode, we refer to the current state as $\textsc{State}(\head)$ and its location as $\textsc{Location}(\head)$. 


\subsection{Find Sections in Multilevel Planning\label{sec:multilevelplanner}}

\begin{algorithm}[t]
\algcaption{MultilevelPlanner($\xi, \xg, \X_1,\ldots,\X_K$)}
\begin{algorithmic}[1]
  \Require Initial state $\xi$, goal state $\xg$, state spaces $\X_1,\ldots,\X_K$
  \Ensure Graphs $G_1,\ldots,G_K$
  \State Let $\PriorityQueue$ be a \Call{priority\ queue}{}\label{alg:bundleplanner:priorityqueue}
  \For{$k=1$ to $K$}
    \def\Xbest{\X_{\text{select}}}
    \def\Xcur{\X_{k}}
    \State $\Call{FindSection}{\Xcur}$\label{alg:bundleplanner:section}
    \State $\PriorityQueue.\Call{push}{\Xcur,
    \textsc{Importance}(\Xcur)}$\label{alg:bundleplanner:pushk}
    \While{$\neg\Call{ptc}{\Xcur}$}\label{alg:bundleplanner:while}
      \State $\Xbest = \PriorityQueue.\Call{pop}{}$\label{alg:bundleplanner:popselect}
      \State $\Call{Grow}{\Xbest}$
      \label{alg:bundleplanner:growselect}
      \State $\PriorityQueue.\Call{push}{\Xbest, \textsc{Importance}(\Xbest)}$\label{alg:bundleplanner:pushselect}
    \EndWhile
  \EndFor
  \State \Return $\Call{Graphs}{\X_1,\ldots,\X_K}$
\end{algorithmic}\label{alg:bundleplanner}
\end{algorithm}

\begin{algorithm}[t]
\algcaption{FindSection($\Xk$)}
\begin{algorithmic}[1]
    \If{\Call{Exists}{$\Xkk $}}\label{alg:section:exist}
    \State $\basePath \gets \Call{BasePath}{\G_{k-1}}$
    \State $\restriction \gets \Call{Restriction}{\basePath}$
    \State $\head \gets \Call{HeadPointer}{x_I, \location=0, \restriction}$
    \State $\Call{PatternDance}{\head}$
    \EndIf
\end{algorithmic}\label{alg:findsection}
\end{algorithm}

The find section problem is a subproblem of the more general multilevel motion planning problem (see Sec.~\ref{sec:background:multilevel}). In previous works, we proposed to solve multilevel planning problems using a dedicated multilevel planner \cite{Orthey2020IJRR}. To clarify the role of finding sections, we describe this multilevel planner in Alg.~\ref{alg:bundleplanner}. We initialize this algorithm with an initial state $\xi$, a goal state $\xg$ and a sequence of bundle spaces $\X_1, \ldots, \X_K$. To search for a feasible path, we first initialize a priority queue (Line 1), then we iteratively explore the bundle spaces (Line 2) by first trying to solve the find section problem (Line 3), then pushing the $k$-th bundle space into the priority queue (Line 4). We compute the importance of a bundle space by the sampling density of its associated graph \cite{Orthey2020IJRR} as
\begin{equation}
\textsc{Importance}(X_k) = \dfrac{1}{|V_k|^{1/n_k}+1}
\end{equation}
with $|V_k|$ being the number of nodes in the graph $G_k$ on $X_k$ and $n_k$ is the dimensionality of $X_k$. We then go into a while loop which terminates if a planner terminate condition (PTC) of the $k$-th space is not fulfilled (Line 5). A PTC can be a timelimit, an iteration limit or a desired cost. We then pop the space with the highest importance from the queue (Line 6), execute one grow iteration for the selected bundle space (Line 7) and push the space back to the queue thereby updating its importance (Line 8). The planner terminates if the PTC of all bundle spaces is false and returns the graphs of all computed levels (Line 11). From those graphs, we can then compute the (optimal) solution path using a discrete A* search \cite{Hart1968} (if one exists). All multilevel planner share this high-level structure. Multilevel planners differ by how the \textsc{Grow} function is implemented. 

We previously developed four multilevel planners. First, the quotient-space roadmap planner (QMP), in which we implement \textsc{Grow} as a probabilistic roadmap (PRM) step \cite{Kavraki1996}. Second, the quotient-space rapidly-exploring random tree (QRRT), in which we implement \textsc{Grow} as an RRT step \cite{Kuffner2000}. Finally, we use the two asymptotically optimal versions QRRT* and QMP*, in which we implement a step of RRT* 
and PRM* \cite{Karaman2011}, respectively. The algorithms also differ in how we compute the distance metric and how we implement sampling inside the grow function, as we detail in our previous publication \cite{Orthey2020IJRR}.

The main contribution of our paper, the pattern dance algorithm, is an efficient method to solve the find section problem. The integration into the multilevel planner is shown in the \textsc{FindSection} method in Alg.~\ref{alg:findsection}. First, we check if there exists a base space (Line 1). We then compute a base path $p$ from the underlying graph or tree on the base space (Line 2). We then build a path restriction $r$ from $p$ (Line 3) and create a head on the path restriction (Line 4). We then call the pattern dance algorithm with the head as input. 

\subsection{Pattern Dance Algorithm}

\def\algName{PatternDance}

\begin{algorithm}[t]
\algcaption{\algName($\head, \depth=0$)}
\begin{algorithmic}[1]
\Params Maximum branching factor $\maxBranch$, maximum depth factor $\maxDepth$, base space step size $\deltaBase$
    \If{$\Call{ManhattanPattern}{\head}$}
        \State \Return \True
    \EndIf
    \If{$\depth \geq \maxDepth$}
    \State \Return \False
    \EndIf
    \If{\Call{WrigglePattern}{\head} \OR \Call{TunnelPattern}{\head}}
    \State \Return \Call{\algName}{\head, \depth+1}
    \EndIf
    \State $l \gets \Call{Location}{\head}+\deltaBase$
    \State $x_B \gets \Call{BasePathAt}{p,l}$
    \For{$j \in [1, \maxBranch]$}
       \State $x_F \gets \Call{SampleFiber}{x_B}$
       \State $x \gets \Call{Lift}{x_B, x_F}$
       \State $\xhead \gets \Call{State}{\head}$
       \If{$\Call{IsValid}{x} \AND
       \neg \Call{CheckMotion}{\xhead, x}$}
       \If{$\Call{TripleStepPattern}{\head, x}$}
           \State \Return \Call{\algName}{\head, \depth+1}
       \EndIf
       \EndIf
    \EndFor
\end{algorithmic}\label{alg:patterndance}
\end{algorithm}

We depict the pseudocode of the pattern dance algorithm in Alg.~\ref{alg:patterndance}. The input is a head over the path restriction and a recursion depth (initially set to zero). Inside the pattern dance algorithm, we coordinate the execution of four section patterns. The rational behind the coordination is to try less complex patterns first while we can successfully move the head forward along the path restriction. Only if no progress is made, we revert to more and more complex patterns to resolve the situation. We found this to be an efficient strategy to quickly find sections.

Those four section patterns are detailed in Sec.~\ref{sec:sectionpatterns} and either move the head forward by controlling the lowest amount of joints possible (\textsc{ManhattanPattern}), execute random walk steps with forward bias (\textsc{WrigglePattern}), try to overcome small barriers using steps outside the path restriction (\textsc{TunnelPattern}) or use a dedicated backtracking procedure (\textsc{TripleStepPattern}) to efficiently find feasible path sections.

Before going into detail, we provide a brief summary and motivation. The algorithm iterates through all four patterns, starting with the computationally cheapest \textsc{ManhattanPattern} (Line 1). If the pattern succeeds, we successfully return (Line 2). Otherwise, we check if we reached the maximum recursion depth (Line 4) and return with failure (Line 5). 

If the depth is below the maximum depth, we continue by executing first the \wrigglePattern and the \tunnelPattern (Line 7). If one pattern successfully terminates, we recursively call the pattern dance algorithm and we increase the recursion depth (Line 8). If no pattern successfully terminates, we backtrack using the \triplestepPattern. To execute the triple step pattern, we first interpolate a single step forward along the base path (Line 10, 11). We then attempt to find a valid fiber space element for a maximum of $\maxBranch$ attempts (Line 12). This is done by first sampling a fiber state over the given base state (Line 13). We then lift the state to the path restriction (Line 14) to obtain a state $x$. If this state is valid and we \emph{cannot} reach it from the head state (Line 16), we execute the triple step pattern with target $x$ (Line 17). If we successfully executed the pattern, we call the pattern dance algorithm again recursively. Note that the small forward step of $\deltaBase$ (Line 10) is an essential component of our algorithm. If we would sample directly over the head base state, we often would sample symmetrical local minima (as an example, see state $p^{\prime}_1$ in Fig.~\ref{fig:sectionpattern:triplestep}). We found this to be particularly important for higher dimensional state spaces, where we often encounter infinitely many symmetrical local minima (consider the set of horizontal rotations of the cylinder before entering the opening in the Bugtrap scenario in Sec.~\ref{sec:evaluations}).  

To implement the section patterns and the pattern dance algorithm, we use the open motion planning library (OMPL) \cite{Sucan2012}. The algorithms are freely available and part of our multilevel motion planning extension of OMPL \cite{Orthey2020IJRR}. All code can be downloaded over github\footnote{\url{https://github.com/aorthey/MotionExplorer/} and \url{https://github.com/aorthey/ompl/}.}.
All parameters used in the algorithms are shown in Table~\ref{tab:parameters}, including the values we use for the evaluations. The values for $\maxBranch, \maxSample, \maxDepth$ are chosen as large as possible to still give good performance on our hardware.

\begin{table}[t]
\centering
\begin{tabular}{ccc}
\toprule
Parameter & Description & Values used\\
\midrule
\maxDepth & Maximum depth of pattern dance & $3$\\
\maxBranch & Maximum branching of pattern dance & $500$\\
\maxSample & Maximum sampling attempts & $100$\\
\deltaBase & Step size on base space & $0.01 \mu_{\Xkk}$\\
\deltaFiber & Step size on fiber space & $0.01 \mu_{\fiberk}$\\
\bottomrule 
\end{tabular}
\caption{Parameters used in algorithm. The variable $\mu_{\X}$ refers to the measure (volume) of the state space $\X$.\label{tab:parameters}}
\end{table}
\section{Section Patterns\label{sec:sectionpatterns}}

The pattern dance algorithm relies on four section patterns, to which we like to provide more detail and motivation. Each of those section patterns is a particular approach to efficiently traverse narrow passages and escape local minima, whereby a local minimum is defined as a region where the heuristic cost is only weakly correlated with the true cost-to-go \cite{Vats2017}. Each section pattern takes as input a head pointer and tries to move this head pointer forward along the path restriction. Please also consult Fig.~\ref{fig:pathrestriction} for visualization of the terminology used.

\subsection{Manhattan Pattern}

Our first section pattern to propagate the head pointer $\head$ is the Manhattan (\ManhattanAbbrv) pattern. With the \ManhattanAbbrv pattern, we interpolate a path between the head state and the goal state along the path restriction. To interpolate, we first interpolate along the base path while keeping the fiber element fixed. Once we reach the end of the base path, we interpolate along the fiber space to the goal state. This method is motivated by our desire to actuate the smallest number of joints at the same time, which is advantageous for high-dimensional systems \cite{Cortes2008}.

\begin{algorithm}[t]
\algcaption{ManhattanPattern($\head$)}
\begin{algorithmic}[1]
\Params Base space step size $\deltaBase$
\State $\xhead \gets \Call{State}{\head}$
\State $x_{F} \gets \Call{ProjectFiber}{\xhead}$
\Comment{$\pif(\xhead)$}
\State $l \gets \Call{Location}{\head}$
\State $s \gets \emptyset$
\While{$l < \Call{Length}{p}$}
    \State $x_B \gets \Call{BasePathAt}{p,l}$ \Comment{State $p(l)$ on base path}
    \State $x \gets \Call{Lift}{x_B, x_{F}}$
    \State $s \gets s \cup \{x\}$
    \State $l \gets l + \deltaBase$
\EndWhile
\State $s \gets s \cup \{x_G\}$
\State $\head \gets \Call{CheckMotion}{s}$\Comment{Return Last Valid}
\State \Return $\Call{HasReachedGoal}{\head}$
\end{algorithmic}\label{alg:L1pattern}
\end{algorithm}

We detail the \ManhattanAbbrv pattern in Alg.~\ref{alg:L1pattern}. We take as input a head pointer $\head$ over a path restriction $r$ with base path $p$. We first project the head state onto the fiber (Line 1-2) by using the fiber projection $\pif$. We then take the location of the head pointer along the base path (Line 3) and step along the base path in increments of $\deltaBase$ (Line 5-10) and add the states to the path $s$ (Line 4). This is done by computing the next base state (Line 6), lifting the base state into the total space (Line 7) and adding it to the path (Line 8). Once we reached the end of the base path, we add the goal state to the section (Line 11). The resulting path $s$ is schematically shown in Fig.~\ref{fig:pathrestriction}. Finally, we evaluate the path by moving along until a constraint violation occurs or we reached the goal state (Line 12). The function $\textsc{CheckMotion}$ returns the last valid state which we use to update the head $\head$. We then return true if the head has reached the goal and false otherwise.

\begin{figure}
    \centering
    \includegraphics[width=0.55\linewidth]{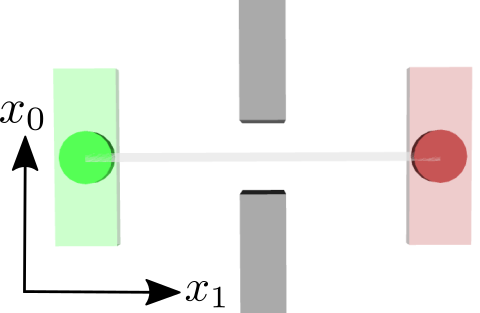}
    \includegraphics[width=0.4\linewidth]{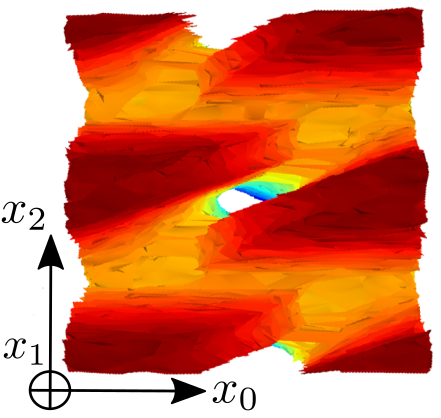}
    \caption{Left: Rectangular rigid robot which has to traverse a narrow passage from a green start to a red goal state. Right: The geometry of its state spaces (darker colors are closer to start state).}
    \label{fig:geometrynarrowpassage}
\end{figure}

\begin{figure}
    \centering
    \includegraphics[width=0.48\linewidth]{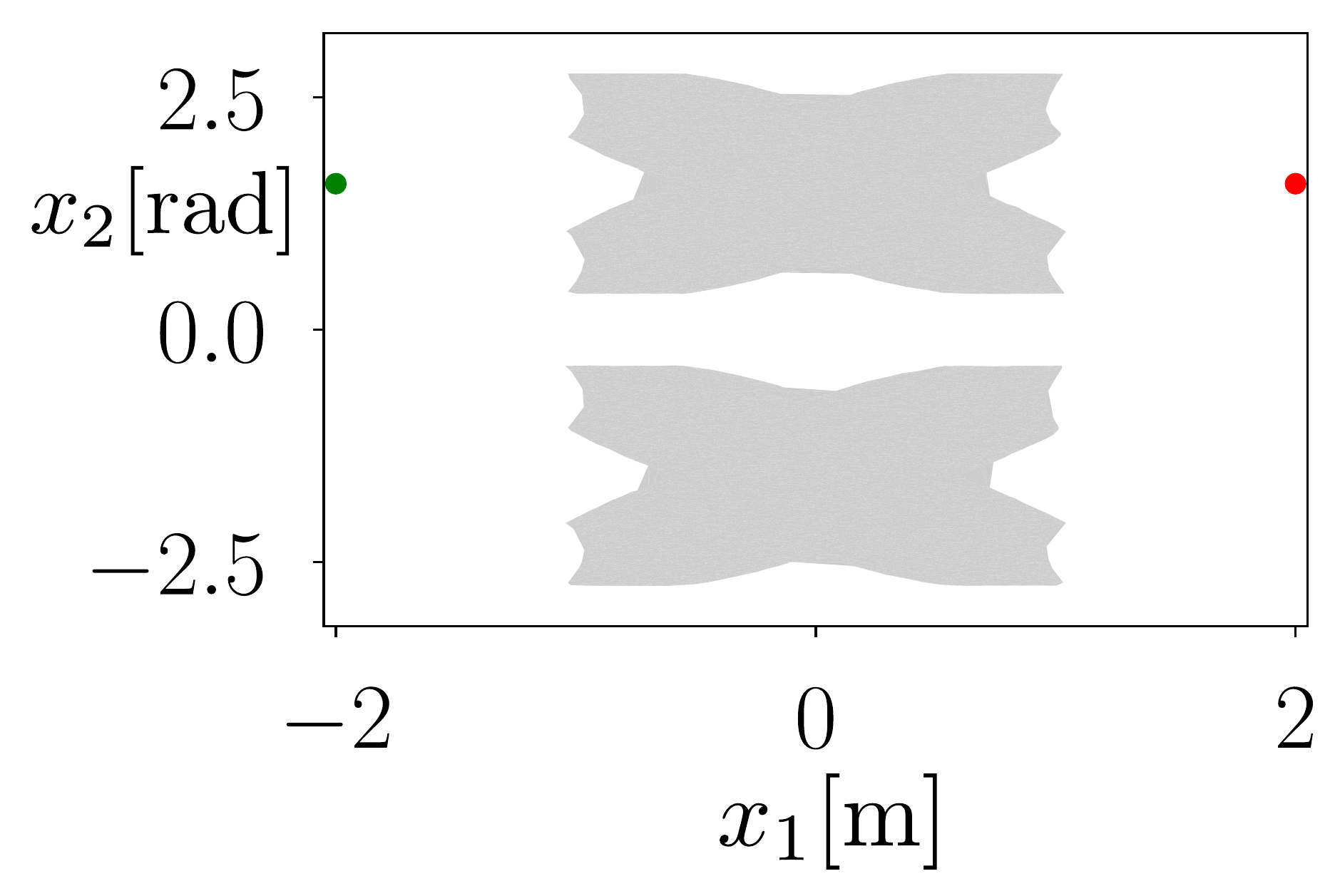}
    \includegraphics[width=0.48\linewidth]{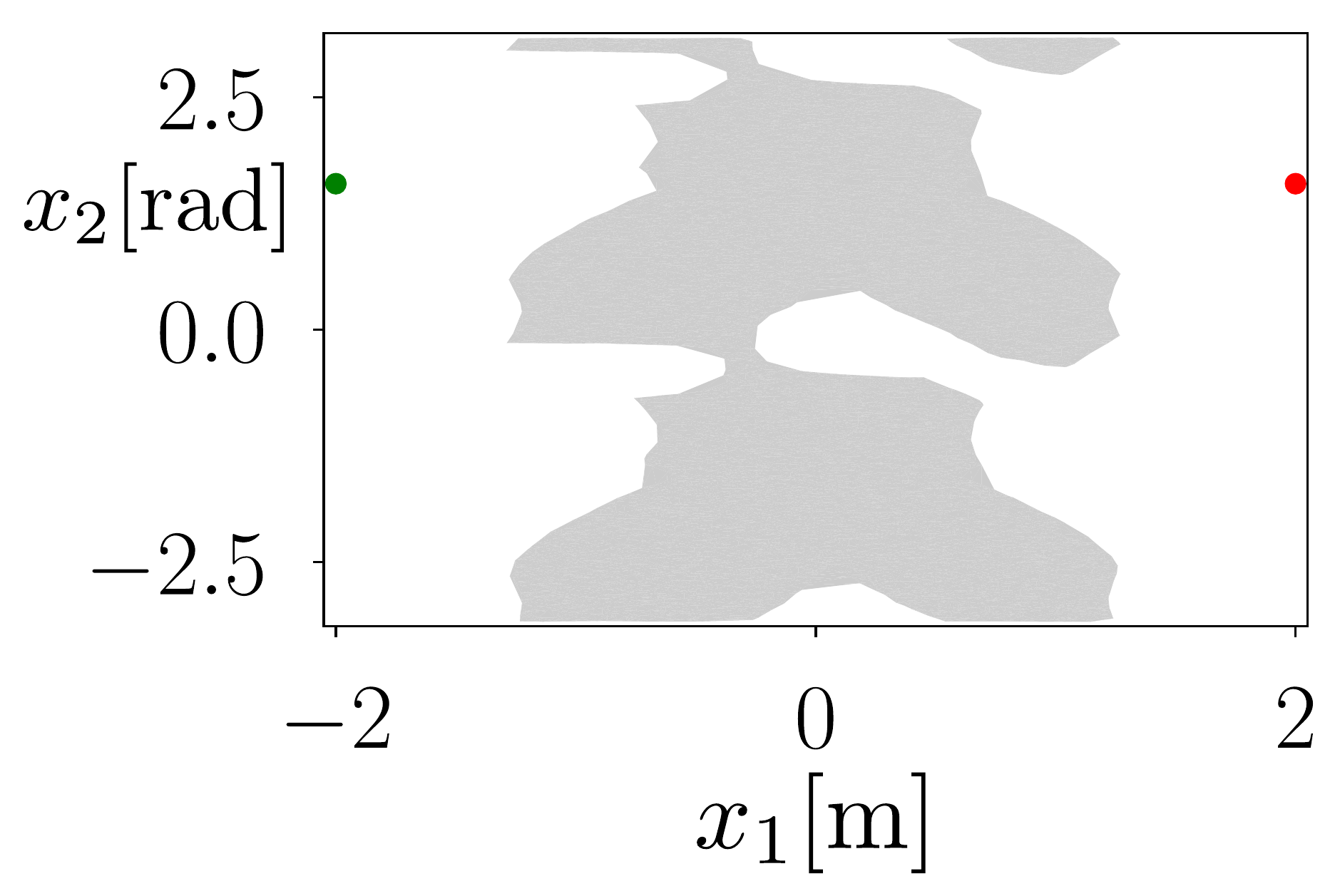}
    \caption{Two path restrictions for the rectangular rigid robot near a narrow passage. Left: Robot moves along a straight base path. Right: Robot moves along a slanted base path.}
    \label{fig:pathrestrictionsnarrowpassages}
\end{figure}

\subsection{Interlude: The Geometry near Narrow Passages}

The next three section patterns are tailor-made solutions to either traverse a narrow passage or to escape a local minimum. To motivate those patterns, we first study the geometry of state spaces near narrow passages. We use a simple toy example of a rigid rectangular body moving in the 2D plane. The state space of this rigid body is the special Euclidean group $SE(2)$, consisting of position and orientation. We assume that the body is located near to a narrow passages as shown in Fig.~\ref{fig:geometrynarrowpassage} (left). We will further assume that our task is to move the rigid body through the narrow passage, from a start state (green) to a goal state (red). We will represent a state as $(x_0,x_1,x_2) \in SE(2)$, with $x_0, x_1$  being vertical and horizontal displacement and $x_2$ the orientation. We visualize a subset of the state space in Fig.~\ref{fig:geometrynarrowpassage} (right), whereby points in collision are colored from dark red (low $x_1$ value, close to start) to bright blue (high $x_1$ value, close to goal). 

To generate path restrictions, we first use a relaxation of the problem onto a circular disk as shown in Fig.~\ref{fig:geometrynarrowpassage} (Left). We model this relaxation using the fiber bundle $SE(2) \rightarrow \R^2$ with base space $\R^2$ and total space $SE(2)$ \cite{Orthey2018}. Let us assume a base path $p: I \rightarrow \R^2$ for the disk to be given. This path induces a two-dimensional path restriction in $SE(2)$, two of which we visualize in Fig.~\ref{fig:pathrestrictionsnarrowpassages}. The left figure shows a path restriction for a base path going straight through the passage, as shown in Fig.~\ref{fig:geometrynarrowpassage}. The right figure shows a path restriction for a base path which goes slanted through the passage. Both are also slices through the state space geometry shown in Fig.~\ref{fig:geometrynarrowpassage} (right). From Fig.~\ref{fig:pathrestrictionsnarrowpassages}, we observe that there are at least three failure cases. Either, we reach a local minimum, we collide with constraints near a narrow passage or we get stuck in front of a small but infeasible region. For each case, we develop a dedicated section pattern to either advance or backtrack.

\subsection{Wriggle Pattern}

\begin{figure}
    \centering
    \includegraphics[width=\linewidth]{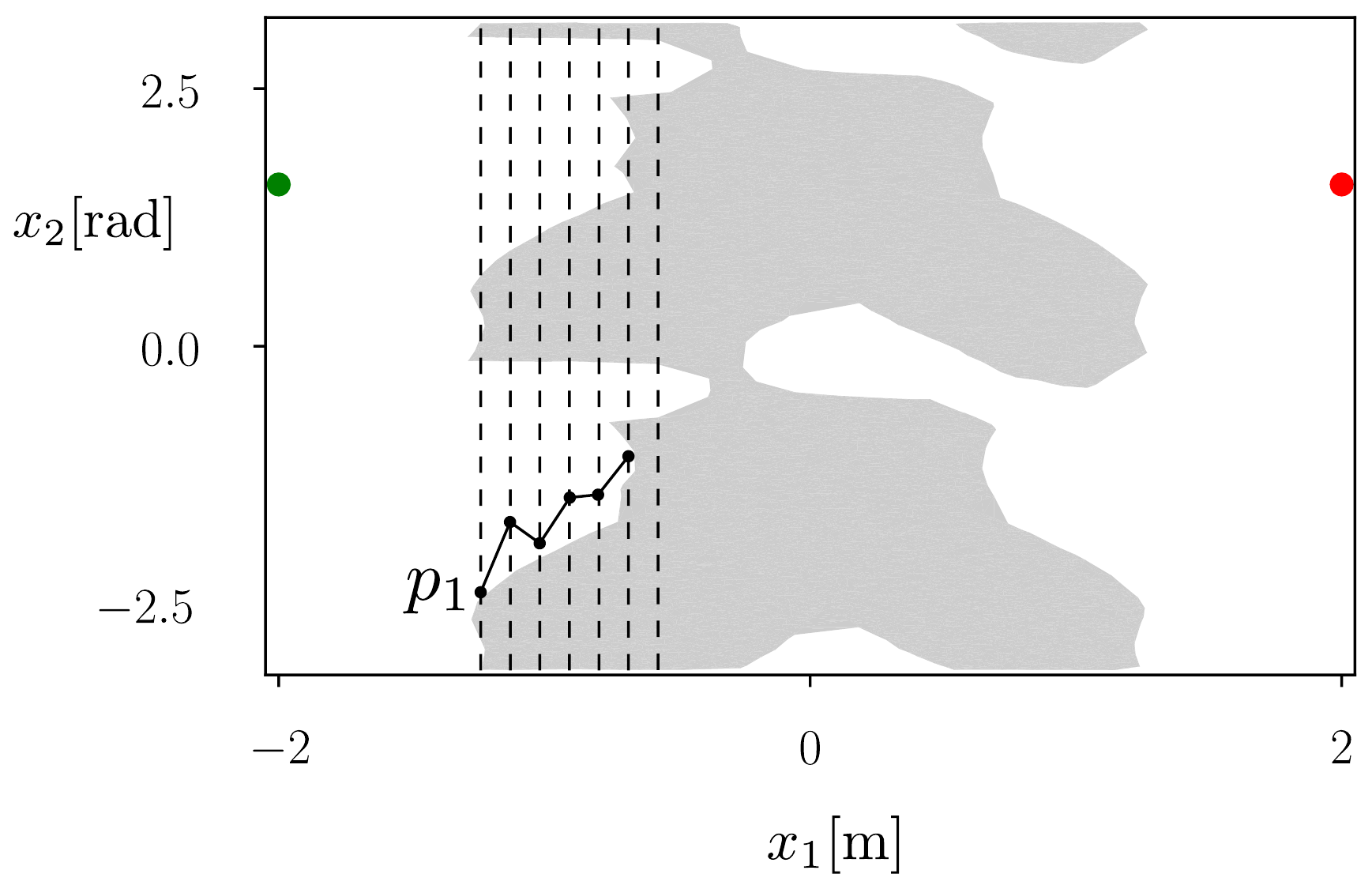}
    \caption{Wriggle pattern to traverse a narrow passage: Given a feasible state $p_1$, we make coordinated random walk steps along the fibers of the path restriction. The distance between fibers is determined by the base space step size parameter $\deltaBase$.\label{fig:wrigglepattern}}
\end{figure}

If we reach a local minimum, the triple step pattern is a way to backtrack to a narrow passage. However, we often might execute the triple step pattern prematurely, because we bumped into constraints near or in a narrow passage. To circumvent those situations, we use the wriggle pattern. With the wriggle pattern, we make coordinated random steps along the fibers of the path restriction and accept a step if it is valid, which is similar to retraction-based sampling \cite{Zhang2008Retraction}. We visualize this pattern in Fig.~\ref{fig:wrigglepattern}.

We show the pseudocode in Alg.~\ref{alg:wrigglepattern}. We start by making one $\deltaBase$ step forward from the head (Line 1). Until we have not reached the end (Line 3), we get the base state at location $l$ (Line 4), and get the fiber element of the head state (Line 6). We then sample for $\maxSample$ rounds (Line 8) by sampling a fiber state in the $\deltaFiber$ proximity of the head fiber state (Line 9). We then lift the base and fiber state (Line 10) and check if the state is valid (Line 11). If the state is valid, we check if the motion from the head to the new state is feasible (Line 12-17). We terminate if we could not expand the state (Line 21-23) or reach the end. We then return true if we made at least one step (Line 25).

\def\counterSampler{\ensuremath{\text{ctr}}}
\def\stepsTaken{\ensuremath{\text{steps}}}

\begin{algorithm}[t]
\algcaption{WrigglePattern($\head$)}
\begin{algorithmic}[1]
\Params Base space step size $\deltaBase$, fiber space step size $\deltaFiber$, maximum samples $\maxSample$
\State $l \gets \Call{Location}{\head} + \deltaBase$
\State $\stepsTaken \gets 0$
\While{$l < \Call{Length}{p}$}
    \State $x_B \gets \Call{BasePathAt}{p,l}$
    \State $\xhead \gets \Call{State}{\head}$
    \State $x_{\fiber_{\head}} \gets \Call{ProjectFiber}{\xhead}$
    \State $\counterSampler \gets 0$
    \While{$\counterSampler < \maxSample$}
        \State $x_F \gets \Call{SampleUniformNear}{x_{\fiber_{\head}}, \deltaFiber}$
        \State $x \gets \Call{Lift}{x_B, x_{F}}$
        \If{$\Call{IsValid}{x}$}
            \If{$\Call{CheckMotion}{\xhead, x}$}
                \State $\G_k \gets \G_k \cup \{\xhead, x\}$
                \State $\Call{UpdateHead}{\head, x}$
                \State $\stepsTaken \gets \stepsTaken + 1$
                \State \BREAK
            \EndIf
        \EndIf
        \State $\counterSampler \gets \counterSampler + 1$
    \EndWhile
    \If{$\counterSampler \geq \maxSample$}
        \State \BREAK
    \EndIf
\EndWhile
\State \Return $\stepsTaken > 0$
\end{algorithmic}\label{alg:wrigglepattern}
\end{algorithm}

\subsection{Tunnel Pattern}

\begin{figure}
    \centering
    \includegraphics[width=\linewidth]{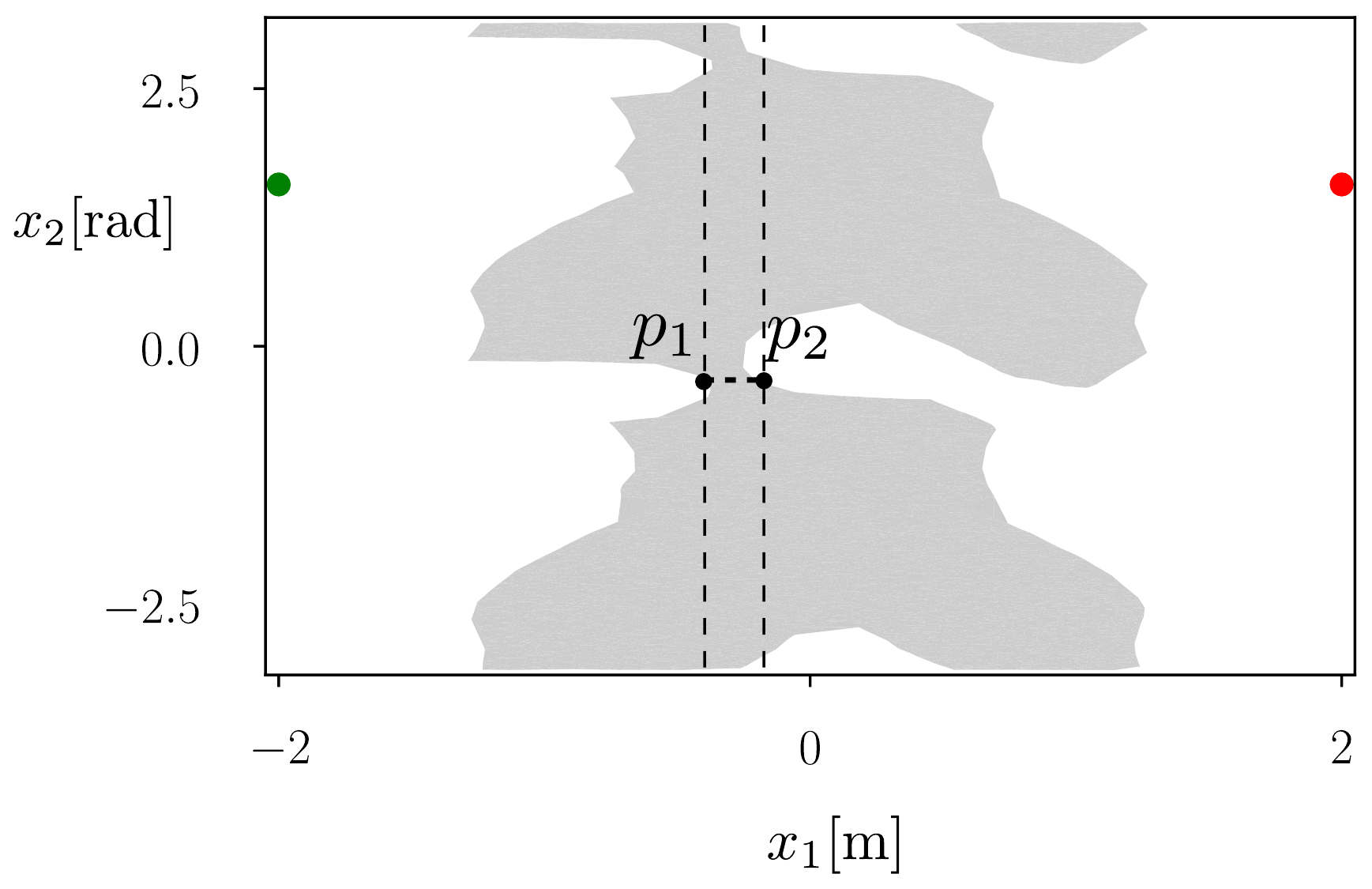}
    \caption{Tunnel pattern to traverse a narrow passage: Given two feasible states $p_1$ and $p_2$, we connect them by momentarily leaving the path restriction to circumnavigate the infeasible region between them.\label{fig:tunnelpattern}}
\end{figure}
\begin{algorithm}[!ht]
\algcaption{TunnelPattern($\head$)}
\begin{algorithmic}[1]
\Params Base space step size $\deltaBase$, fiber space step size $\deltaFiber$, maximum samples $\maxSample$

\State $(\xend, \lend) \gets \Call{TunnelEnd}{\head}$
\State $\xhead \gets \Call{State}{\head}$
\State $\xfiberhead \gets \Call{ProjectFiber}{\xhead}$

\def\dbest{d_{\text{best}}}
\State $\dbest \gets \Call{Distance}{\xhead, \xend}$
\State $l \gets \Call{Location}{\head}$
\While{$l \leq \lend$}
    \If{$\Call{CheckMotion}{\xhead, \xend}$}
        \State $\G_k \gets \G_k \cup \{\xhead, \xend\}$
        \State $\Call{UpdateHead}{\head, \xend}$
        \State \Return \True
    \EndIf
    \State $l \gets l + \deltaBase$
    \State $x_B \gets \Call{BasePathAt}{p,l}$
    \State $\epsilon \gets \Call{SmoothParameter}{0,10\deltaBase,\maxSample}$
    \State $\counterSampler \gets 0$
    \While{$\counterSampler < \maxSample$}
        \State $x_B \gets \Call{SampleUniformNear}{x_B, \epsilon(\counterSampler)}$
        \State $x_\fiber \gets \Call{SampleUniformNear}{\xfiberhead, \deltaFiber}$
        \State $x \gets \Call{Lift}{x_B, x_\fiber}$
        \If{$\Call{IsValid}{x}$}
            \State $d \gets \Call{Distance}{x, \xend}$
            \If{$d < \dbest \AND 
            \Call{CheckMotion}{\xhead, x}$}
                \State $\G_k \gets \G_k \cup \{\xhead, x\}$
                \State $\xhead \gets x$
                \State \BREAK
            \EndIf
        \EndIf
        \State $\counterSampler \gets \counterSampler + 1$
    \EndWhile
    \If{$\counterSampler \geq \maxSample$}
    \State \Return \False
    \EndIf
\EndWhile
\State \Return \False
\end{algorithmic}\label{alg:tunnelpattern}
\end{algorithm}

While the wriggle pattern locally explores the neighborhood \emph{inside} the path restriction, we often encounter situations where we find it advantageous to momentarily step \emph{outside} the path restriction to overcome an infeasible region. From the perspective of the path restriction, we ``tunnel'' through the infeasible region, which we therefore refer to as the tunnel pattern. With the tunnel pattern, we assume to be located at a local minimum $p_1$ as shown in Fig.~\ref{fig:tunnelpattern}. To resolve this situation, we try to find the next valid state $p_2$ while keeping the fiber element constant. We then try to connect $p_1$ to $p_2$ by sampling valid states in a smoothly increasing neighborhood of the base space and a constant neighborhood in fiber space. While $p_2$ is not reached, we accept new states if they decrease the distance to $p_2$.

We show the pseudocode in Alg.~\ref{alg:tunnelpattern}. We first search for a tunnel ending state $\xend$ at base path location $\lend$ (Line 1). To find the tunnel ending, we step forward along the base path without changing the fiber until we find a valid state. We then try to connect the head state $\xhead$ to the tunnel ending state $\xend$. We use a while loop to move along the relevant base path segment from the head location $l$ to the tunnel end location $\lend$ (Line 6). We first check if we can connect the head state to the tunnel end state (Line 7). If true, we add a new edge into the graph (Line 8), set the head to the tunnel ending state (Line 9) and return true (Line 10). Otherwise, we step forward along the base path with step size $\deltaBase$ (Line 12) and query the base state at $l$ (Line 13). Instead of using the base state exactly, we use a smoothly increasing neighborhood parameter $\epsilon$. The value of $\epsilon$ depends on the counter $\counterSampler$ and smoothly interpolates between $0$ and $10 \deltaBase$ using an Hermite polynomial \cite{De1987} (Line 14). We then attempt to make a step towards the tunnel ending for a maximum of $\maxSample$ attempts (Line 16). We do this by sampling a base space element (Line 17) and a fiber element (Line 18). We then lift the state (Line 19) and check for validity (Line 20). If the new state is valid, its distance is closer to the tunnel ending and we can connect it to the head state (Line 22), we add a new edge to the graph (Line 23), set the head state to the new state (Line 24) and continue forward (Line 25). If we fail to find a better sample for $\maxSample$ attempts, we return false (Line 30-32). We also return false if we reach the base path location $\lend$ without having a valid connection (Line 34).

\subsection{Triple Step Pattern}
\begin{figure}[t]
    \centering
    \begin{subfigure}[t]{\linewidth}
    \centering
    \includegraphics[width=\linewidth]{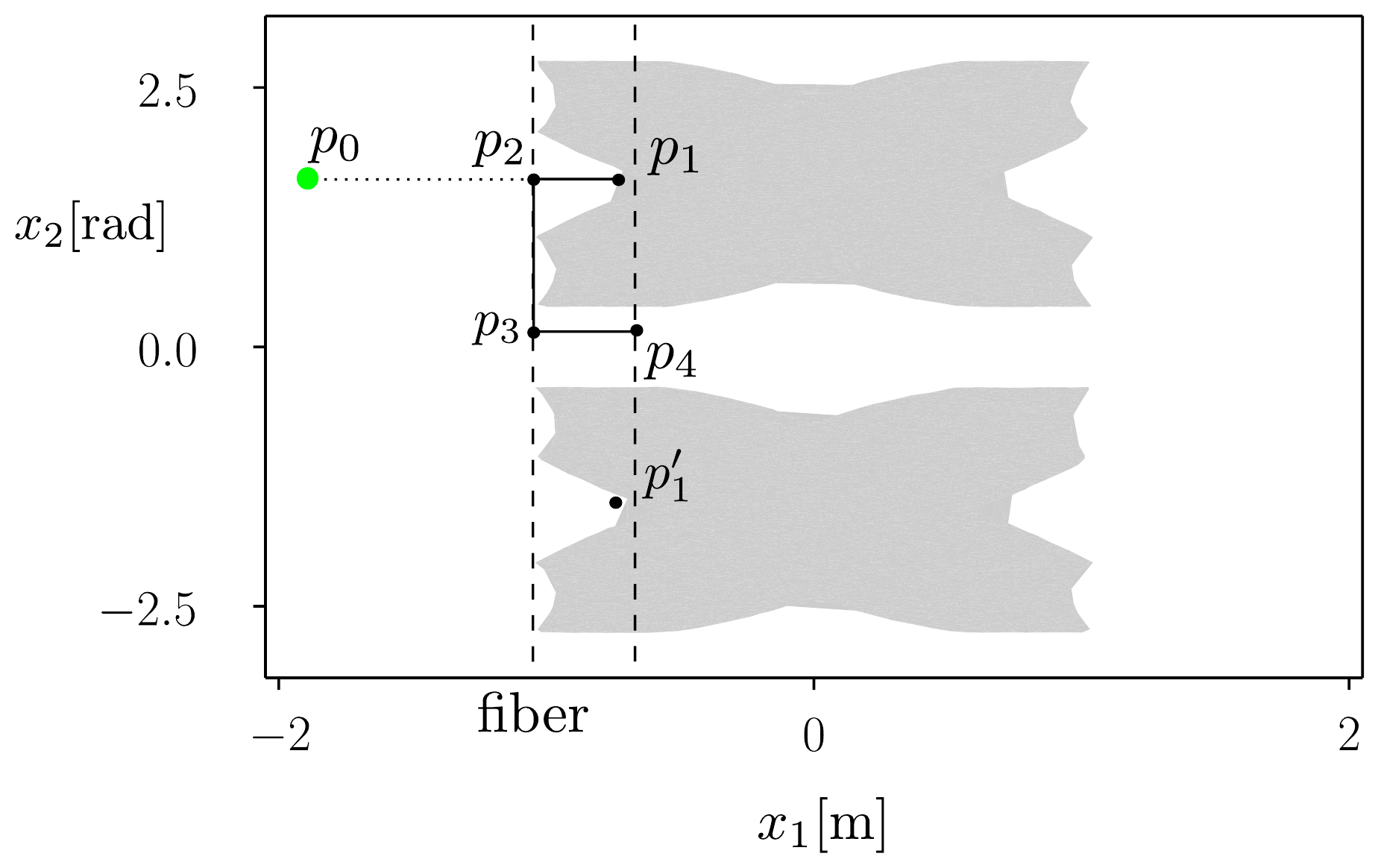}
    \end{subfigure}
    \begin{subfigure}[t]{0.23\linewidth}
    \centering
    \includegraphics[width=\linewidth]{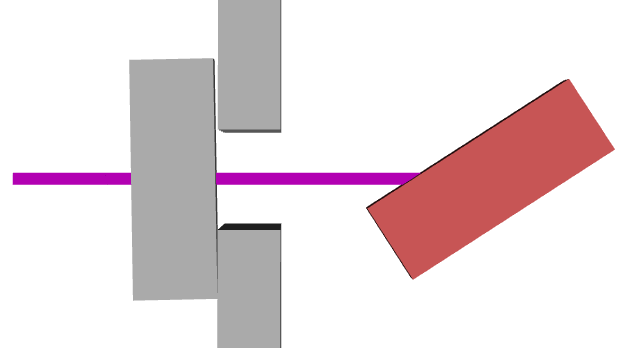}
    \caption{At $p_1$ (after collision).}
    \end{subfigure}
    \begin{subfigure}[t]{0.24\linewidth}
    \centering
    \includegraphics[width=\linewidth]{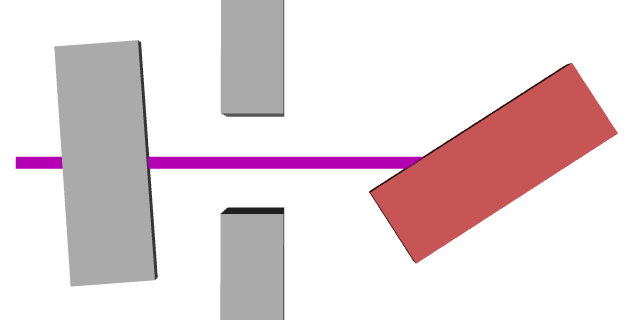}
    \caption{At $p_2$ (after backstep).}
    \end{subfigure}
    \begin{subfigure}[t]{0.24\linewidth}
    \centering
    \includegraphics[width=\linewidth]{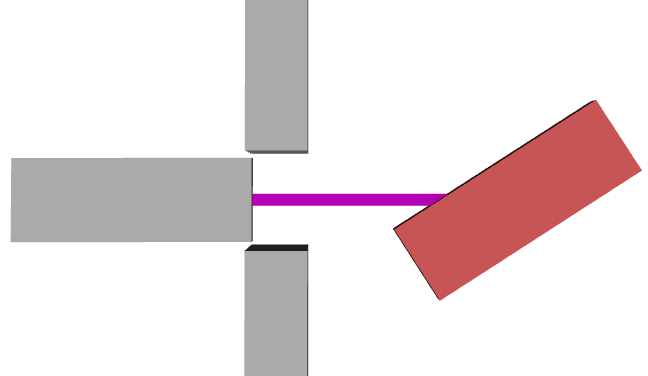}
    \caption{At $p_3$ (after sidestep).}
    \end{subfigure}
    \begin{subfigure}[t]{0.24\linewidth}
    \centering
    \includegraphics[width=\linewidth]{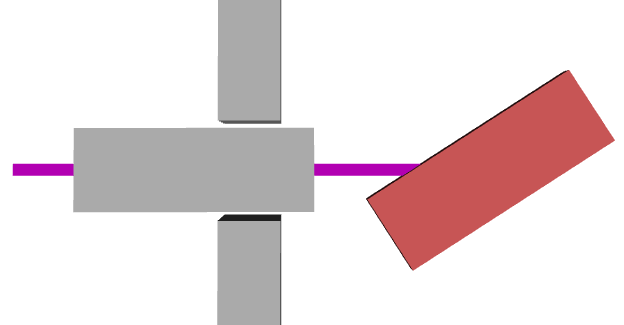}
    \caption{At $p_4$ (after forward step).}
    \end{subfigure}

\caption{Triple step pattern to traverse a narrow passage: We start at a state $p_1$ (a), backstep to a state $p_2$ (b), sidestep along the fiber to $p_3$ (c) and then step forward to reach a state $p_4$ (d). \label{fig:sectionpattern:triplestep}}
\end{figure}
\begin{algorithm}[t]
\algcaption{TripleStepPattern($\head, x$)}
\begin{algorithmic}[1]
\Params Base space step size $\deltaBase$
\State $\xhead \gets \Call{State}{\head}$
\State $l \gets \Call{Location}{\head}$
\State $x_{F_1} \gets \Call{ProjectFiber}{\head}$
\State $x_{F_2} \gets \Call{ProjectFiber}{x}$
\State $x_{F_m} \gets \Call{Steer}{x_{F_1},x_{F_2}, 0.5}$\Comment{Midpoint}
\While{$l > \deltaBase$}
    \State $l \gets l - \deltaBase$
    \State $x_B \gets \Call{BasePathAt}{p,l}$
    \State $\xm \gets \Call{Lift}{x_B, x_{F_m}}$
    \If{$\Call{IsValid}{\xm}$}
        \State $x_1 \gets \Call{Lift}{x_B, x_{F_1}}$
        \State $x_2 \gets \Call{Lift}{x_B, x_{F_2}}$
        \If{$\Call{CheckMotion}{x_1,x_2}$}
            \If{$\Call{CheckMotion}{\xhead, x_1}$}
                \If{$\Call{CheckMotion}{x_2, x}$}
                    \State $\G_k \gets \G_k \cup \{\xhead, x_1\}$
                    \State $\G_k \gets \G_k \cup \{x_1, x_2\}$
                    \State $\G_k \gets \G_k \cup \{x_2, x\}$
                    \State $\Call{UpdateHead}{h, x}$
                    \State \Return \True
                \EndIf
            \EndIf
            \State \BREAK \Comment{End While Loop}
        \EndIf
    \EndIf
\EndWhile
\State \Return \False
\end{algorithmic}\label{alg:triplesteppattern}
\end{algorithm}

To escape a local minimum, we develop the triple step pattern. With the triple step pattern, we connect two states on the path restriction using a triple backtracking step. 

The idea of the triple step pattern is to connect two states on (or near) the same fiber. Before explaining the pattern in detail, we first visualize the pattern in  Fig.~\ref{fig:sectionpattern:triplestep}. You can see a rectangular rigid body in the plane, which is currently at state $p_1$ (Fig.~\ref{fig:sectionpattern:triplestep} (a)) and which we like to move to state $p_4$ (Fig.~\ref{fig:sectionpattern:triplestep} (d)). To connect those states, we first move backwards along the path restriction from $p_1$ to another state $p_2$ (Fig.~\ref{fig:sectionpattern:triplestep} (b)) while moving from $p_4$ to another state $p_3$ (Fig.~\ref{fig:sectionpattern:triplestep} (c)), respectively. We move backwards until we can connect $p_2$ and $p_3$ by a straight line segment. In that case we execute a backstep from $p_1$ to $p_2$, a sidestep (along the fiber marked) from $p_2$ to $p_3$ and a forward step from $p_3$ to $p_4$. Note that $p_4$ is slightly moved forward such that we avoid situations where we backtrack to a symmetric local minimum like $p'_1$ which would not improve our location along the path restriction.

We show the pseudocode for the triple step pattern in Alg.~\ref{alg:triplesteppattern}. Our goal is to connect the head state to the given state $x$. We first compute a midpoint on the fiber space (Line 5) (to minimize the number of \textsc{CheckMotion} calls \cite{Mandalika2019}). We then move backwards along the base path while we are greater than the parameter $\deltaBase$ (Line 6-7). For each location, we interpolate a base state (Line 8), lift the state using the fiber midpoint (Line 9) and check if this state is valid. If it is valid, we compute intermediate states $x_1$ and $x_2$ (Line 11, 12) and check if the motion between them is feasible (Line 13). If that is true, we additionally check if the backward and forward steps are feasible (Line 14, 15). If that is true, we add those edges to the graph (Line 16-18) and update the head to our new state $x$ (Line 19). In that case we return true (Line 20). If we fail to find such a triple step, we terminate once we reach the beginning of the base path location and return false (Line 27). 

\section{Evaluations\label{sec:evaluations}}

\begin{figure*}[ht]
    \centering
    \begin{subfigure}[t]{0.3\linewidth}
    \centering
    \includegraphics[width=0.9\linewidth]{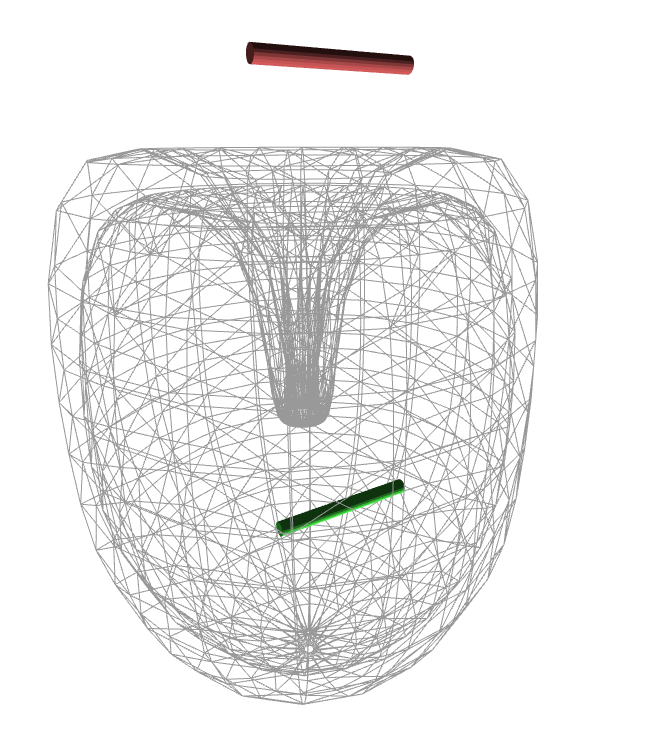}
    \caption{06D Bugtrap\label{fig:scenarios:bugtrap}} 
    \end{subfigure}
    \begin{subfigure}[t]{0.3\linewidth}
    \centering
    \includegraphics[width=0.9\linewidth]{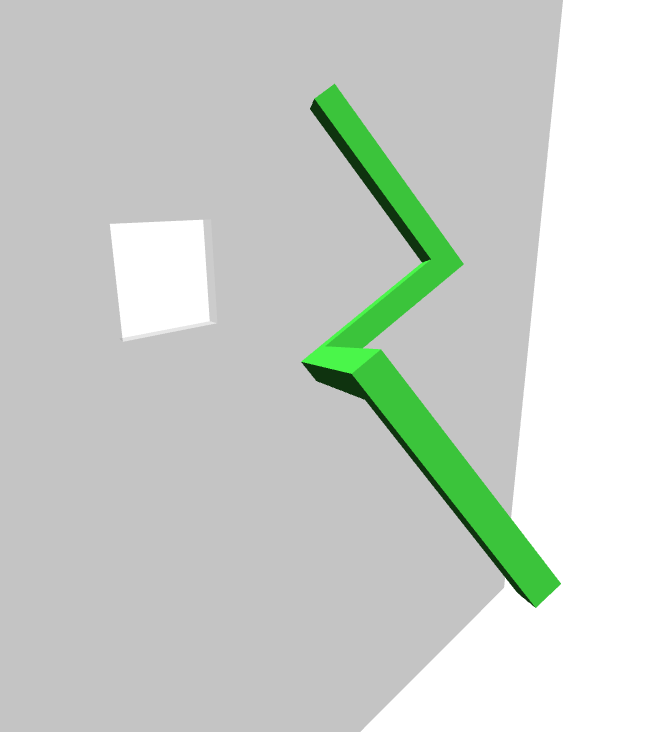}
    \caption{06D Double Lshape\label{fig:scenarios:doubleLshape}} 
    \end{subfigure}
    \begin{subfigure}[t]{0.3\linewidth}
    \centering
    \includegraphics[width=0.9\linewidth]{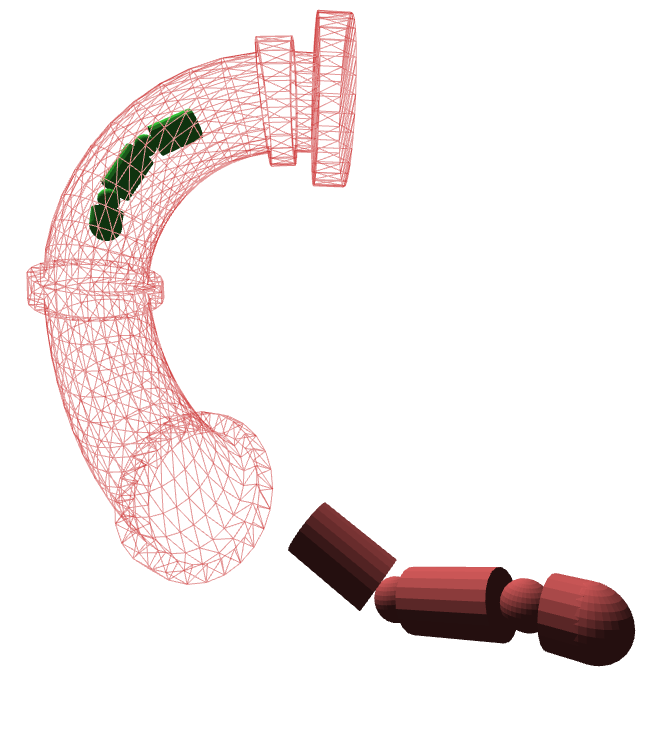}
    \caption{10D Chain Egress\label{fig:scenarios:chainegress}} 
    \end{subfigure}
    \begin{subfigure}[t]{0.25\linewidth}
    \centering
    \includegraphics[width=\linewidth]{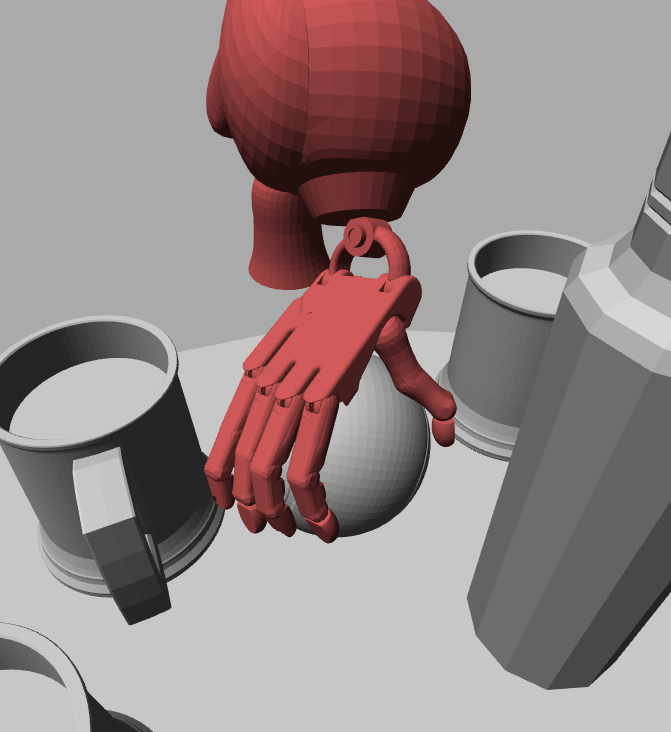}
    \caption{37D ShadowHand Ball\label{fig:scenarios:overhand}} 
    \end{subfigure}
    \begin{subfigure}[t]{0.25\linewidth}
    \centering
    \includegraphics[width=\linewidth]{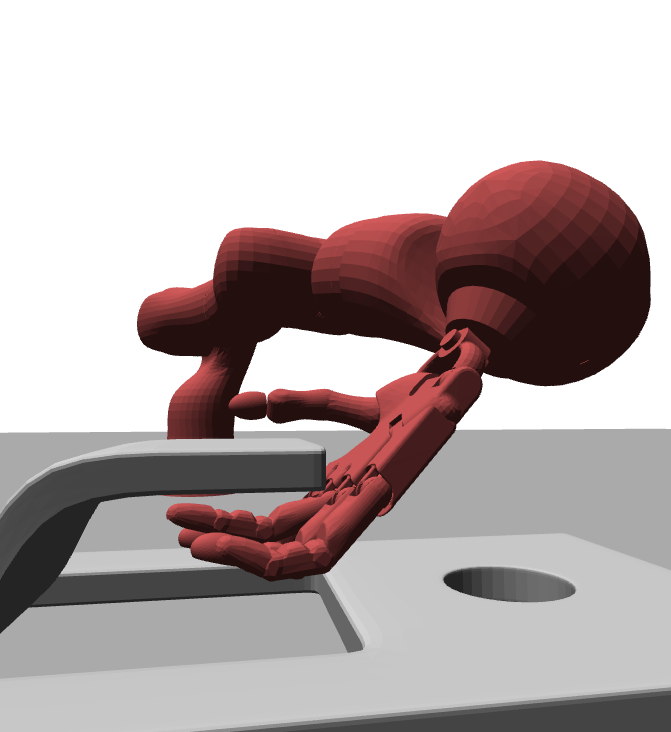}
    \caption{37D ShadowHand Metal\label{fig:scenarios:underhand}} 
    \end{subfigure}
    \begin{subfigure}[t]{0.24\linewidth}
    \centering
    \includegraphics[width=\linewidth]{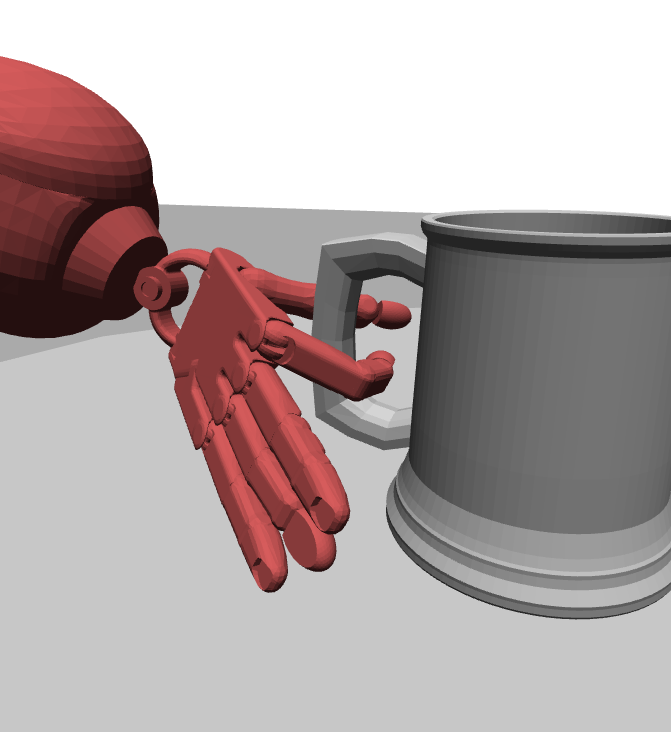}
    \caption{37D ShadowHand Mug\label{fig:scenarios:mug}} 
    \end{subfigure}
    \begin{subfigure}[t]{0.24\linewidth}
    \centering
    \includegraphics[width=\linewidth]{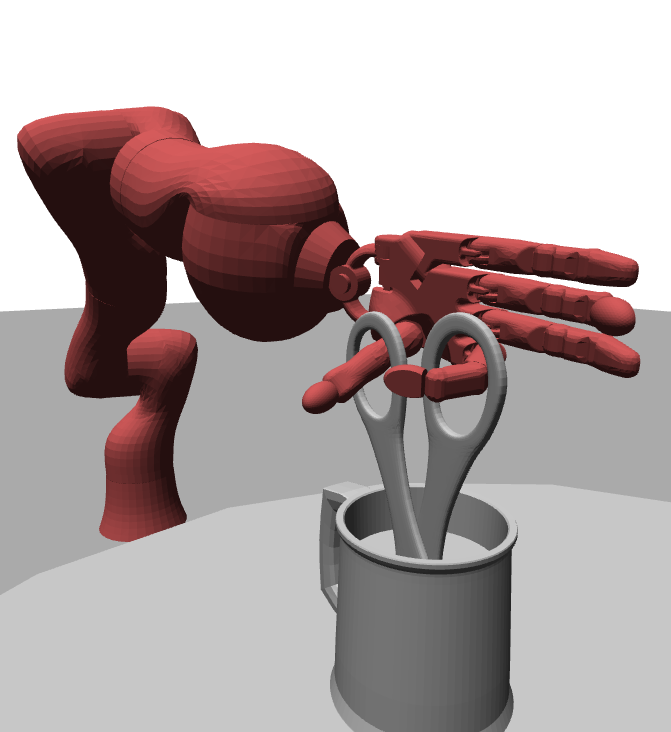}
    \caption{37D ShadowHand Scissor\label{fig:scenarios:scissor}} 
    \end{subfigure}
\caption{Scenarios for evaluations. The task is to move the robot from the start state (green) to the goal state (red). Top Row (left to right): Bugtrap (6-dof), Double L Shape (6-dof) (goal configuration not shown) and Chain Egress (10-dof). Bottom Row: Overhand, Underhand, Single-Finger and Double-Finger Pregrasp (each 37-dof) (start configurations not shown).\label{fig:scenarios}}
\end{figure*}
\begin{figure}[ht]
    \centering
    \begin{subfigure}[t]{0.32\linewidth}
    \centering
    \includegraphics[width=\linewidth]{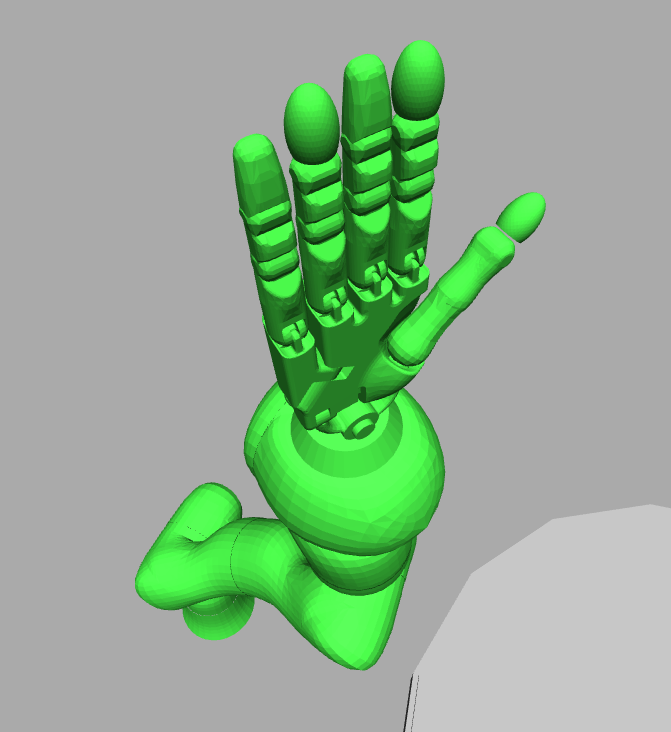}
    \caption{Shadow Hand Level 3 $\R^{37}$.\label{fig:simplifications:hand3}}
    \end{subfigure}
    \begin{subfigure}[t]{0.32\linewidth}
    \centering
    \includegraphics[width=\linewidth]{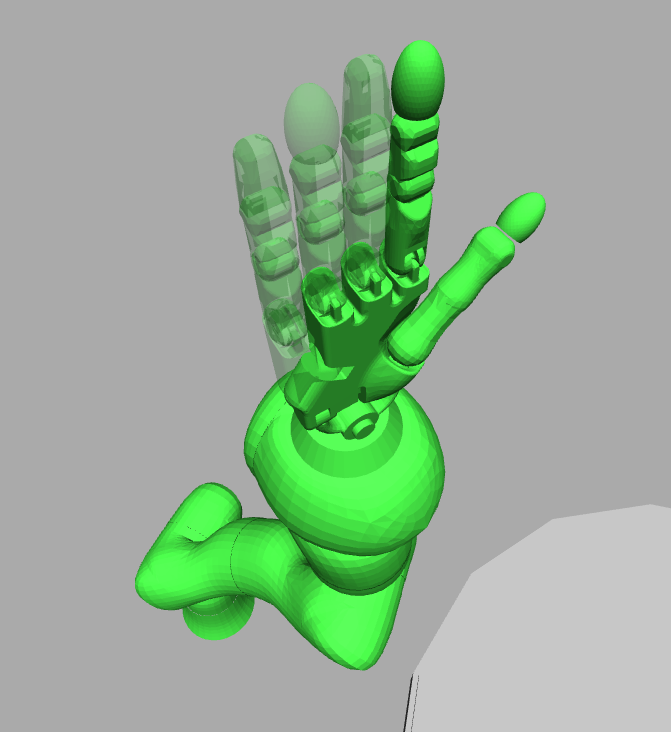}
    \caption{Shadow Hand Level 2 $\R^{18}$.\label{fig:simplifications:hand2}}
    \end{subfigure}
    \begin{subfigure}[t]{0.32\linewidth}
    \centering
    \includegraphics[width=\linewidth]{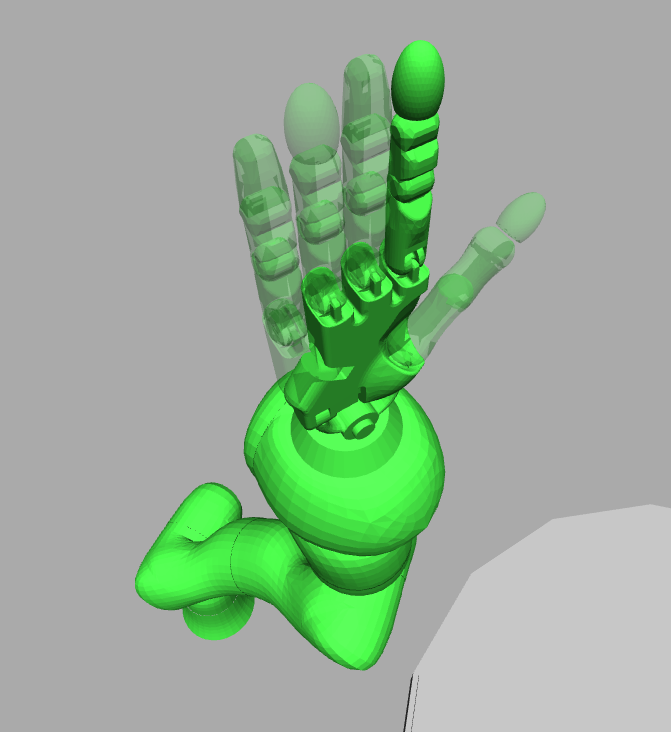}
    \caption{Shadow Hand Level 1 $\R^{13}$.\label{fig:simplifications:hand1}}
    \end{subfigure}
    \begin{subfigure}[t]{0.32\linewidth}
    \centering
    \includegraphics[width=\linewidth]{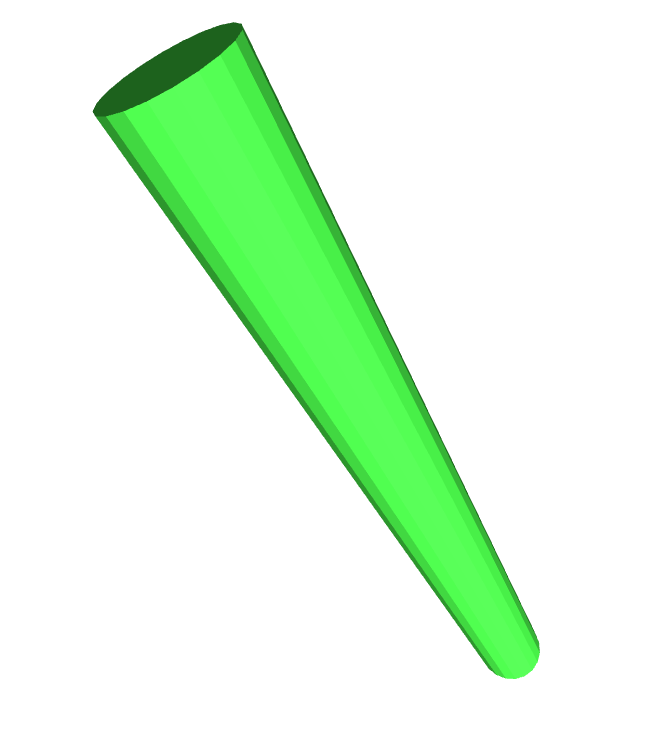}
    \caption{Bugtrap Level 2 $SE(3)$.\label{fig:simplifications:bugtrap2}}
    \end{subfigure}    
    \begin{subfigure}[t]{0.32\linewidth}
    \centering
    \includegraphics[width=\linewidth]{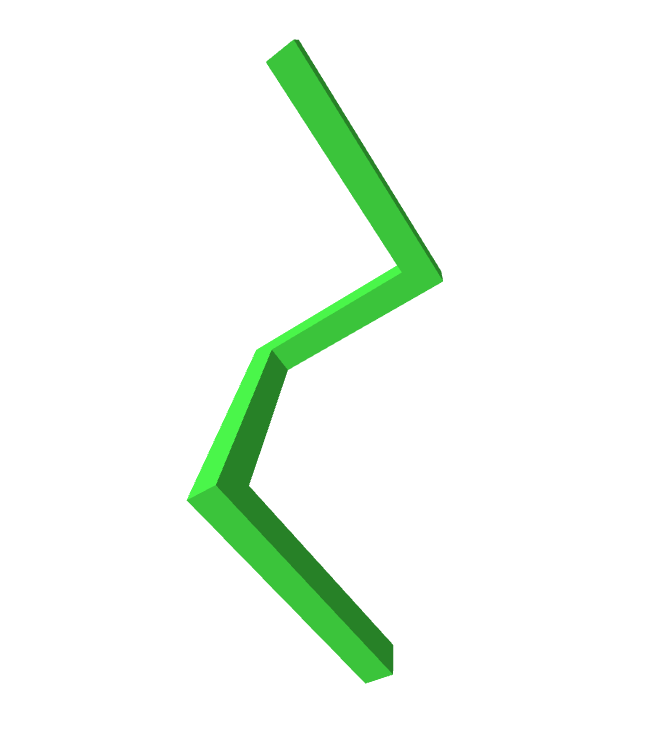}
    \caption{Double Lshape Level 2 $SE(3)$.\label{fig:simplifications:dls2}}
    \end{subfigure}    
    \begin{subfigure}[t]{0.32\linewidth}
    \centering
    \includegraphics[width=\linewidth]{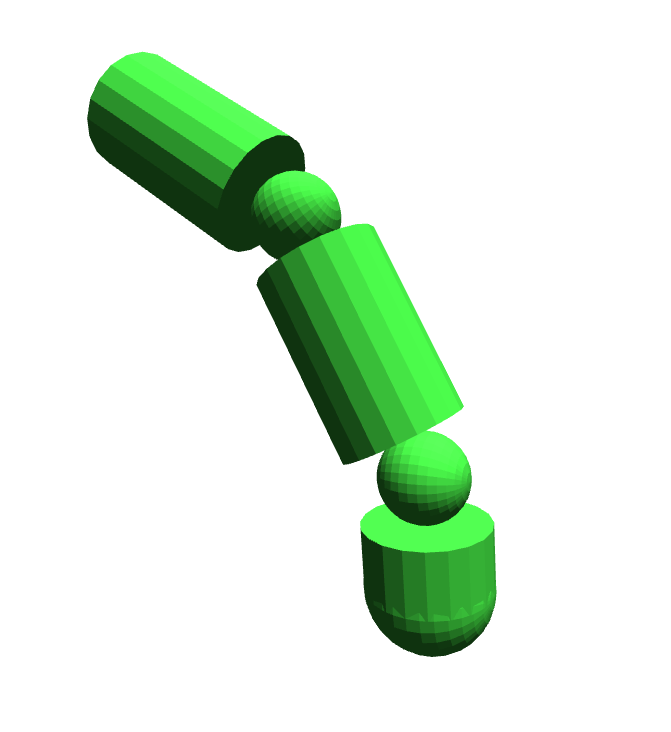}
    \caption{Articulated Chain Level 2 $SE(3) \times \R^6$.\label{fig:simplifications:chain2}}
    \end{subfigure}    
    \begin{subfigure}[t]{0.32\linewidth}
    \centering
    \includegraphics[width=\linewidth]{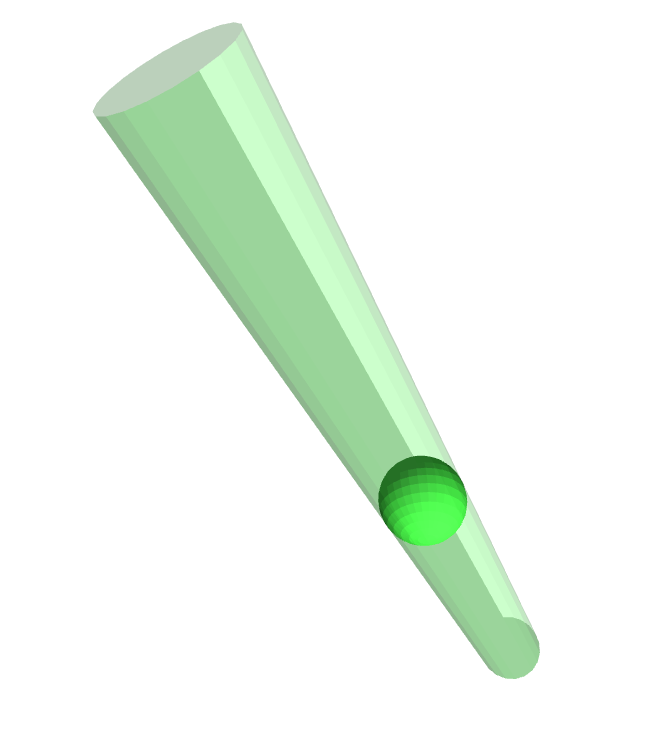}
    \caption{Bugtrap Level 1 $\R^3$.\label{fig:simplifications:bugtrap1}}
    \end{subfigure}       
    \begin{subfigure}[t]{0.32\linewidth}
    \centering
    \includegraphics[width=\linewidth]{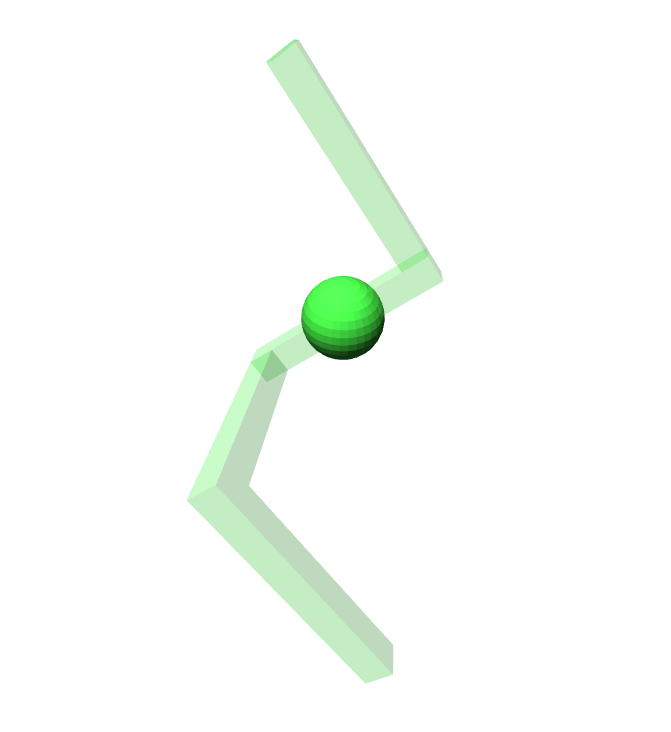}
    \caption{Double Lshape Level 1 $\R^3$.\label{fig:simplifications:dls1}}
    \end{subfigure}    
    \begin{subfigure}[t]{0.32\linewidth}
    \centering
    \includegraphics[width=\linewidth]{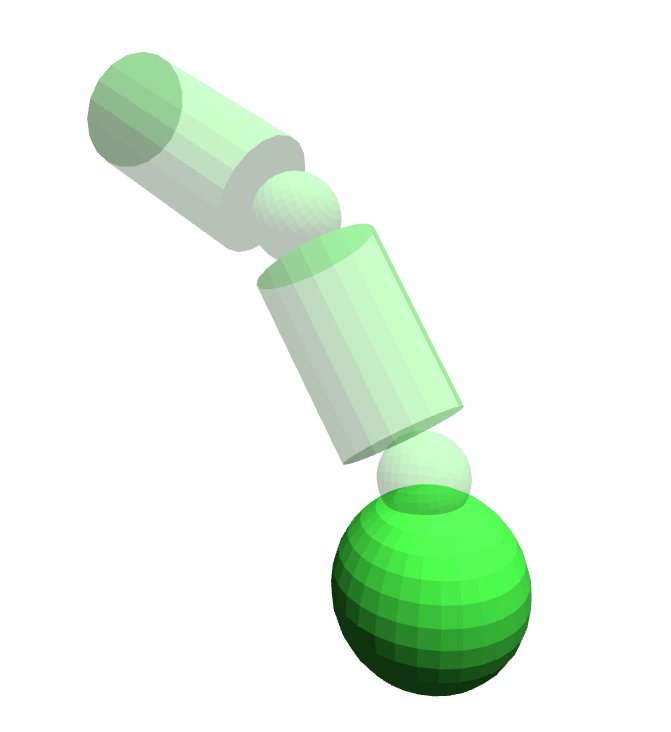}
    \caption{Articulated Chain Level 1 $\R^3$.\label{fig:simplifications:chain1}}
    \end{subfigure}
\caption{Multilevel abstraction using simplified models.\label{fig:simplifications}}
\end{figure}
\colorlet{Mycolor1}{gray!80}
\newcommand{\markedcell}[1]{\textit{\textcolor{Mycolor1}{#1}}}
\newcommand{\thispaper}{{(\textbf{ours})}}
\begin{table}[!t]
\centering
\renewcommand{\cellrotangle}{90}
\renewcommand\theadfont{\bfseries}
\settowidth{\rotheadsize}{\theadfont 37D ShadowHand Scissor}
\newcolumntype{Y}{>{\raggedleft\arraybackslash}X}
\footnotesize\centering
\renewcommand{\arraystretch}{1.2}
\setlength\tabcolsep{3pt}
\begin{tabulary}{\linewidth}{@{}LLCCCCCCC@{}}
\toprule
\multicolumn{2}{>{\centering}p{2.6cm}}{Runtime in seconds (10 run average)} & \rothead{06D Bugtrap} & \rothead{06D Double Lshape} & \rothead{10D Chain Egress} & \rothead{37D ShadowHand Ball} & \rothead{37D ShadowHand Metal} & \rothead{37D ShadowHand Mug} & \rothead{37D ShadowHand Scissor} \\ 
\midrule
1 & \mbox{QRRT \thispaper} &  4.45  &  1.86  &  \textbf{0.55}  &  2.01  &  35.63  &  19.80  &  60.00  \\
2 & \mbox{QRRT* \thispaper} &  24.87  &  2.00  &  0.56  &  25.35  &  43.95  &  60.00  &  60.00  \\ 
3 & \mbox{QMP \thispaper} &  \textbf{0.51}  &  \textbf{1.27}  &  1.91  &  0.86  &  18.98  &  \textbf{1.20}  &  \textbf{14.52}  \\ 
4 & \mbox{QMP* \thispaper} &  0.90  &  1.63  &  7.29  &  \textbf{0.86}  &  \textbf{1.94}  &  1.63  &  37.27  \\ 
\midrule
5 & \mbox{RRT} &  60.00  &  60.00  &  49.77  &  60.00  &  60.00  &  60.00  &  60.00  \\ 
  6 & \mbox{RRTConnect} &  60.00  &  60.00  &  60.00  &  \markedcell{1.70}  &
  \markedcell{8.16}  &  57.38  &  60.00  \\ 
7 & \mbox{RRT\#} &  60.00  &  60.00  &  45.43  &  60.00  &  60.00  &  60.00  &  60.00  \\ 
8 & \mbox{RRT*} &  60.00  &  60.00  &  51.74  &  60.00  &  60.00  &  60.00  &  60.00  \\ 
9 & \mbox{RRTXstatic} &  60.00  &  60.00  &  50.49  &  60.00  &  60.00  &  60.00  &  60.00  \\ 
10 & \mbox{LazyRRT} &  60.00  &  60.00  &  55.56  &  60.00  &  60.00  &  60.00  &  60.00  \\ 
  11 & \mbox{TRRT} &  60.00  &  60.00  &  \markedcell{0.81}  &  42.08  &  60.00  &  60.00  &  60.00  \\ 
  12 & \mbox{BiTRRT} &  11.54  &  54.30  &  \markedcell{4.57}  &  60.00  &  60.00  &  60.00  &  60.00  \\ 
13 & \mbox{LBTRRT} &  60.00  &  60.00  &  60.00  &  60.00  &  60.00  &  60.00  &  60.00  \\ 
  14 & \mbox{RLRT} &  60.00  &  60.00  &  51.39  &  \markedcell{3.68}  &  28.47  &  60.00  &  60.00  \\ 
  15 & \mbox{BiRLRT} &  60.00  &  57.40  &  60.00  &  \markedcell{1.52}  &  25.60  &  60.00  &  60.00  \\ 
16 & \mbox{pRRT} &  60.00  &  60.00  &  49.41  &  60.00  &  60.00  &  60.00  &  60.00  \\ 
17 & \mbox{FMT} &  60.00  &  60.00  &  60.00  &  60.00  &  60.00  &  60.00  &  60.00  \\ 
18 & \mbox{BFMT} &  60.00  &  50.34  &  60.00  &  60.00  &  60.00  &  60.00  &  60.00  \\ 
19 & \mbox{PRM} &  60.00  &  56.47  &  60.00  &  37.25  &  52.72  &  60.00  &  60.00  \\ 
20 & \mbox{PRM*} &  60.00  &  57.80  &  60.00  &  34.24  &  50.04  &  60.00  &  60.00  \\ 
21 & \mbox{LazyPRM} &  60.00  &  60.00  &  60.00  &  60.00  &  60.00  &  60.00  &  60.00  \\ 
22 & \mbox{LazyPRM*} &  60.00  &  60.00  &  60.00  &  54.06  &  60.00  &  60.00  &  60.00  \\ 
23 & \mbox{SPARS} &  60.00  &  59.73  &  60.00  &  60.00  &  60.00  &  60.00  &  60.00  \\ 
24 & \mbox{SPARStwo} &  60.00  &  54.69  &  60.00  &  60.00  &  60.00  &  60.00  &  60.00  \\ 
25 & \mbox{SST} &  60.00  &  60.00  &  60.00  &  60.00  &  60.00  &  60.00  &  60.00  \\ 
26 & \mbox{EST} &  60.00  &  60.00  &  50.46  &  24.96  &  45.64  &  60.00  &  60.00  \\ 
27 & \mbox{BiEST} &  60.00  &  60.00  &  59.85  &  29.79  &  33.36  &  60.00  &  60.00  \\ 
28 & \mbox{InformedRRT*} &  60.00  &  60.00  &  -  &  60.00  &  60.00  &  60.00  &  60.00  \\ 
29 & \mbox{SORRT*} &  60.00  &  60.00  &  -  &  60.00  &  60.00  &  60.00  &  60.00  \\ 
30 & \mbox{kBIT*} &  60.00  &  60.00  &  -  &  34.17  &  46.44  &  60.00  &  60.00  \\ 
31 & \mbox{kABIT*} &  60.00  &  60.00  &  -  &  50.28  &  44.56  &  60.00  &  60.00  \\ 
32 & \mbox{AIT*} &  60.00  &  60.00  &  -  &  55.35  &  60.00  &  60.00  &  60.00  \\ 
33 & \mbox{STRIDE} &  60.00  &  60.00  &  -  &  29.58  &  48.98  &  60.00  &  60.00  \\ 
34 & \mbox{ProjEST} &  60.00  &  60.00  &  -  &  47.77  &  60.00  &  60.00  &  60.00  \\ 
  35 & \mbox{PDST} &  60.00  &  60.00  &  -  &  \markedcell{3.25}  &  54.42  &  60.00  &  60.00  \\ 
  36 & \mbox{KPIECE1} &  60.00  &  60.00  &  -  &  \markedcell{6.27}  &  32.48  &  60.00  &  60.00  \\ 
37 & \mbox{BKPIECE1} &  60.00  &  60.00  &  -  &  52.35  &  60.00  &  60.00  &  60.00  \\ 
38 & \mbox{LBKPIECE1} &  60.00  &  49.79  &  -  &  60.00  &  60.00  &  60.00  &  60.00  \\ 
39 & \mbox{SBL} &  60.00  &  50.30  &  -  &  60.00  &  60.00  &  60.00  &  60.00  \\ 
40 & \mbox{CForest} &  60.00  &  60.00  &  -  &  60.00  &  60.00  &  60.00  &  60.00  \\
\bottomrule
\end{tabulary}
\caption{Runtime (s) of motion planner on the scenarios from Fig. \ref{fig:scenarios}, each         averaged
  over $10$ runs with cut-off time limit of $60$s. An entry $-$ means that
  planner does not support the particular state space.\label{table:eval1}}
\end{table}

\begin{table}[!t]
\centering
\renewcommand{\cellrotangle}{90}
\renewcommand\theadfont{\bfseries}
\settowidth{\rotheadsize}{\theadfont 37D ShadowHand Scissor}
\newcolumntype{Y}{>{\raggedleft\arraybackslash}X}
\footnotesize\centering
\renewcommand{\arraystretch}{1.2}
\setlength\tabcolsep{3pt}
\begin{tabulary}{\linewidth}{@{}LLCCCCCCC@{}}
\toprule
\multicolumn{2}{>{\centering}p{2.8cm}}{Runtime in seconds (10 run average)} & \rothead{06D Bugtrap} & \rothead{06D Double Lshape} & \rothead{10D Chain Egress} & \rothead{37D ShadowHand Ball} & \rothead{37D ShadowHand Metal} & \rothead{37D ShadowHand Mug} & \rothead{37D ShadowHand Scissor} \\ 
\midrule
1 & \mbox{QMP \thispaper} &  \textbf{0.51}  &  \textbf{1.27}  &  \textbf{1.91}  &  \textbf{0.86}  &  \textbf{18.98}  &  \textbf{1.20}  &  \textbf{14.52}  \\ 
2 & \mbox{QMP (SideStepping)} &  60.00  &  26.08  &  60.00  &  1.07  &  55.37  &  6\textsuperscript{a}   &  60.00  \\ 
\midrule
3 & \mbox{QMP* \thispaper} &  \textbf{0.90}  &  \textbf{1.63}  &  \textbf{7.29}  &  \textbf{0.86}  &  \textbf{1.94}  &  \textbf{1.63}  &  \textbf{37.27}  \\ 
4 & \mbox{QMP* (SideStepping)} &  60.00  &  30.11  &  60.00  &  1.76  &  60.00  &  12\textsuperscript{a}   &  60.00  \\ 
\midrule
5 & \mbox{QRRT \thispaper} &  \textbf{4.45}  &  \textbf{1.86}  &  \textbf{0.55}  &  \textbf{2.01}  &  \textbf{35.63}  &  \textbf{19.80}  &  60.00  \\ 
6 & \mbox{QRRT (SideStepping)} &  60.00  &  27.72  &  9.14  &  18.65  &  60.00  &  44\textsuperscript{a}   &  60.00  \\ 
\midrule
7 & \mbox{QRRT* \thispaper} &  \textbf{24.87}  &  \textbf{2.00}  &  \textbf{0.56}  &  \textbf{25.35}  &  \textbf{43.95}  &  60.00  &  60.00  \\ 
8 & \mbox{QRRT* (SideStepping)} &  60.00  &  60.00  &  16.42  &  42.33  &  54.05  &  \textbf{48}\textsuperscript{a}  &  60.00  \\ 
\bottomrule
\end{tabulary}
\caption{Comparison of multilevel planners with sidestepping \cite{Orthey2020IJRR} versus multilevel planner with our pattern dance algorithm.\label{table:eval2}}

\small\textsuperscript{a} Taken from \cite{Orthey2020IJRR}.
\end{table}

To evaluate our pattern dance algorithm, we
integrate it into the multilevel planner QRRT, QRRT*, QMP and QMP*, as we discussed in Sec. \ref{sec:multilevelplanner}. We then conduct two comparisons. First, we compare our planner to $36$
available planning algorithms in the Open motion planning library (OMPL)
\cite{Moll2015} on $7$ challenging environments as shown in Fig.~\ref{fig:scenarios}. For each algorithm, we use the abbreviated name. For a full list of algorithms with full names and associated publication, see \cite{Orthey2020IJRR} and the OMPL documentation \cite{Sucan2012}. Second, we compare the multilevel planner with the pattern dance algorithm to an older version of the same multilevel planner, where we use a recursive
sidestepping algorithm to quickly find sections \cite{Orthey2020IJRR}. 

\subsection{Evaluation Metric}

To evaluate, we use a 8GB RAM 4-core 2.5GHz laptop running Ubuntu 16.04. For each
experiment, we use a minimum length cost (for planner which support cost functions) and we let each planner run $10$ times with a
cut-off time limit of $60$ seconds. We then report on the average runtime over
those $10$ runs. We show the results in Table \ref{table:eval1}. 

Concerning the results, there are two notes of caution. First, we let each OMPL planner run out-of-the-box without any parameter tuning. Further tuning of parameters could potentially improve results significantly. Second, due to the high number of planner and scenarios, we let each planner run only $10$ times and take the average. However, averaging over $10$ runs might exhibit more variance and thereby create more outliers.

\subsection{06-dof Bugtrap}

For the first evaluation, we use the Bugtrap scenario \cite{Lee2012} (Fig.~\ref{fig:scenarios:bugtrap}). The lowest runtime we found in the literature is $22.17$s for a version of the Selective-Retraction-RRT \cite{Lee2012, Zhang2008Retraction}. 
However, this runtime is not directly comparable due to different hardware, implementation, parameters and operating systems. To relax the problem, we use an inscribed sphere at the center of the cylindrical bug as shown in Fig.~\ref{fig:simplifications:bugtrap1} and Fig.~\ref{fig:simplifications:bugtrap2}. 

We show the results of our evaluation in Fig.~\ref{table:eval1}. The best performing planner is QMP (3rd planner in table) with $0.51$s followed by QMP* (4) with $0.90$s and QRRT (1) with $4.45$s. We also see good performance of the BiTRRT (13) planner \cite{Jaillet2010} with $11.54$s. We note that the QRRT* (2) algorithm requires $24.87$s, which we believe to be caused by the additional burden of rewiring the tree \cite{Salzman2016, Orthey2020IJRR}.

\subsection{06-dof Double L shape}

In the next evaluation, we like to show that the section patterns are not specific to the cylindrical geometry, but are more widely applicable to other rigid bodies. As demonstration, we use the double L-shape scenario \cite{VanDenBerg2005}, where two L-shape bodies are connected to each other as shown in Fig.~\ref{fig:scenarios:doubleLshape}. The task is to move through a vertical wall with a small quadratic hole. We use a two-level relaxation by using an inscribed sphere as shown in Fig.~\ref{fig:simplifications:dls1} and Fig.~\ref{fig:simplifications:dls2}. To make our method more robust against base paths too close to obstacles, we increase the size of the sphere slightly to increase clearance from obstacles. 

Our evaluation shows that QMP performs best with $1.27$s followed by QMP* ($1.63$s), QRRT ($1.86$s) and QRRT* ($2.00$s). The next best planner from OMPL is LBKPIECE1 (38) with $49.79$s. 

\subsection{10-dof Chain Egress}

In the third evaluation, we like to increase the complexity by considering an articulated chain ($10$-dof) as shown in Fig.~\ref{fig:scenarios:chainegress}. The task is to remove the chain from a pipe, a typical egress scenario. Note that for such systems, we can find analytical feasible path sections if we assume the base path of the head to be curvature constrained \cite{Orthey2018RAS}. 
However, we will not make such assumption in this paper. 

To relax the problem, we use an inscribed sphere in the head of the chain as shown in Fig.~\ref{fig:simplifications:chain1} and Fig.~\ref{fig:simplifications:chain2}. As in the case of the double L-shape, we slightly increase the size of the sphere to make our method more robust against base paths too close to obstacles. 

In our evaluations, we show that QRRT performs best with $0.55$s followed by QRRT* ($0.56$s). The next best planners are TRRT (11) ($0.81$s), QMP ($1.91$), BiTRRT (12) ($4.57$s) and QMP* with $7.29$s. Note that there are $12$ OMPL planner which cannot address this problem, because they do not support compound state spaces or do not have dedicated projection functions for such spaces.

\subsection{37-dof Pre-Grasp}

For the next evaluations, we compute (pre-)grasping paths for a ShadowHand mounted on a KUKA LWR robot. The tasks are to compute an overhand grasp on a ball (Fig.~\ref{fig:scenarios:overhand}), an underhand grasp on a metal piece 
(Fig.~\ref{fig:scenarios:underhand}), a single-finger precision grasp on a mug (Fig.~\ref{fig:scenarios:mug}) and a double-finger precision grasp on a scissor (Fig.~\ref{fig:scenarios:scissor}). The starting state for all scenarios is an upright position of the arm with hand being open, as shown in Fig.~\ref{fig:simplifications:hand3}. To relax the problem, we use a three-level abstraction by first removing three fingers (Fig.~\ref{fig:simplifications:hand2}) and subsequently removing the thumb (Fig.~\ref{fig:simplifications:hand1}) of the hand. 

Our evaluations show the following results. First, for the Ball scenario, we see that QMP and QMP* perform best with $0.86$s. The next best planner is the OMPL planner BiRLRT (15) \cite{Luna2020} with $1.52$s, QRRT with $2.01$s and RRTConnect (6) with $1.70$s. We note that also the planner PDST (35) \cite{Ladd2004}, RLRT (14) \cite{Luna2020} and KPIECE1 (36) \cite{Sucan2011} perform competively with $3.25$s, $3.68$s and $6.27$s, respectively. The planner QRRT* does not perform well on this problem instance with $25.35$s, due to similar problems as on the Bugtrap scenario. Second, for the underhand grasp on the metal piece, we see that QMP* performs best with $1.94$s followed by RRTConnect (6) with $8.16$s and QMP with $18.98$s. We will address the discrepancy between QMP and QMP* further in Sec.~\ref{sec:limitations}. Third, for the single-finger precision grasp on the mug, we observe that QMP performs best with $1.20$s followed by QMP* with $1.63$s. While QRRT performs significantly worse ($19.80$s), QRRT* was not able to solve this problem ($60.00$s). Fourth, for the double-finger precision grasp on the scissor, we observe that QMP performs best with $14.52$s followed by QMP* with $37.27$s. No other planner is able to solve this problem. We will further discuss the high runtime of both QMP and QMP* in detail in Sec.~\ref{sec:limitations}.
\section{Limitations and Discussion\label{sec:limitations}}

While our evaluations support the usage of section patterns for narrow passage planning problems, we also like to point out two limitations of our approach. To each limitation, we will discuss possible ways to eventually address and resolve the limitation. 

\begin{figure}[ht]
    \centering
    \begin{subfigure}[t]{0.46\linewidth}
    \centering
    \includegraphics[width=\linewidth]{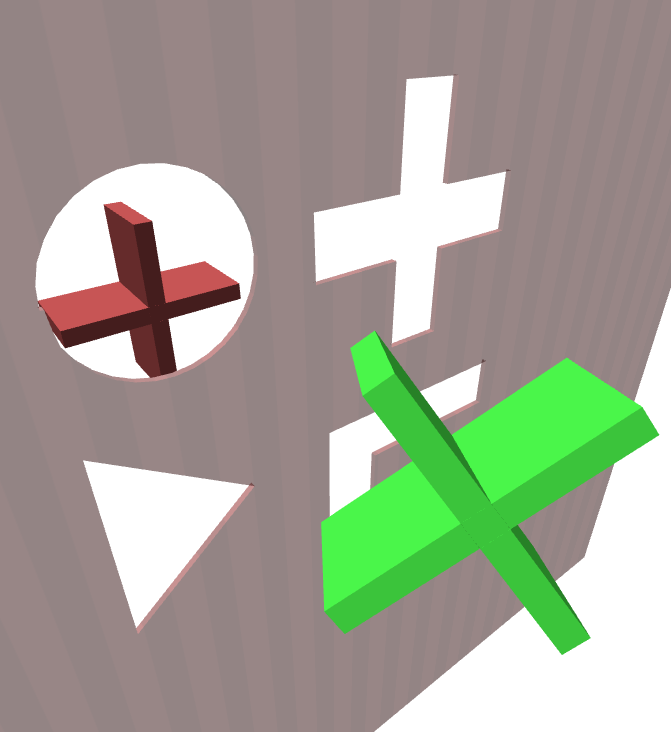}
    \end{subfigure}
    \begin{subfigure}[t]{0.46\linewidth}
    \centering
    \includegraphics[width=\linewidth]{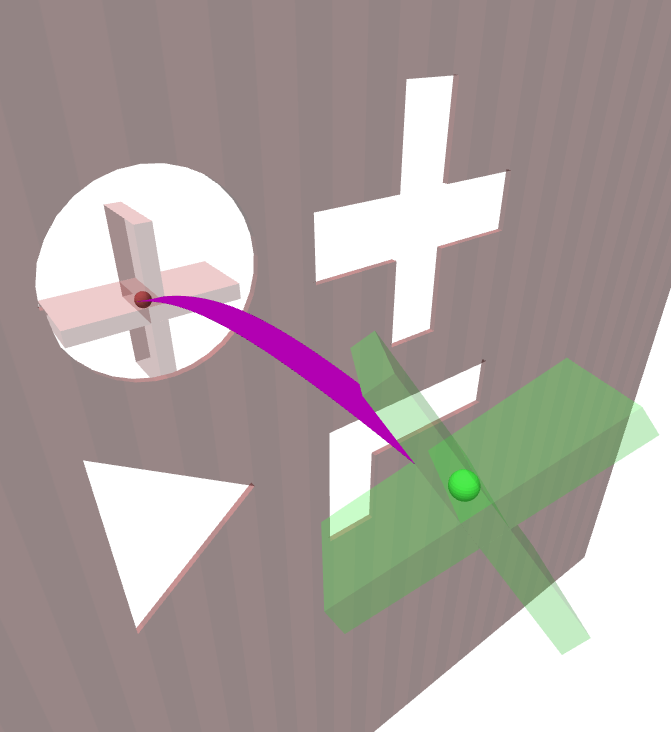}
    \end{subfigure}
\caption{Limitations of Section Pattern Approach. Base path does not admit a feasible path section. See text for clarification. \label{fig:limitations}}
\end{figure}

\begin{table}[!t]
\centering
\renewcommand{\cellrotangle}{90}
\renewcommand\theadfont{\bfseries}
\newcolumntype{Y}{>{\raggedleft\arraybackslash}X}
\footnotesize\centering
\renewcommand{\arraystretch}{1.2}
\setlength\tabcolsep{3pt}
\begin{tabulary}{\linewidth}{@{}LCCCCCCCCCC@{}}
\toprule
 Run & 1 & 2 & 3 & 4 & 5 & 6 & 7 & 8 & 9 & 10\\ 
\midrule
\multicolumn{11}{c}{37D ShadowHand Metal Scenario}\\ 
\midrule
 QMP & 1.53 & 1.11 & 1.20 & 0.99 & 1.06 & 60.00 & 60.00 & 2.93 & 1.02 & 60.00 \\ 
 QMP* & 0.98 & 1.15 & 0.93 & 1.23 & 2.73 & 1.13 & 1.03 & 7.61 & 0.98 & 1.65\\
\midrule
 \multicolumn{11}{c}{37D ShadowHand Scissor Scenario}\\
\midrule
 QMP & 1.45 & 1.50 & 2.14 & 2.17 & 60.00 & 60.00 & 2.44 & 7.49 & 1.51 & 6.51 \\ 
 QMP* & 60.00 & 60.00 & 2.22 & 60.00 & 6.27 & 60.00 & 60.00 & 60.00 & 1.92 & 2.30 \\ 
\bottomrule
\end{tabulary}
\caption{Runtime (s) for QMP and QMP* on each run. Average runtimes are $18.98$s/$1.94$s (QMP/QMP*) for the Metal scenarios and $14.52$s/$37.27$s for the Scissor scenario.\label{table:limitations}}
\end{table}

\subsection{Increased runtime on Metal and Scissor Scenario}

The first limitation is the increased runtime of our planner on the 37D ShadowHand Scissor and the Metal scenario. We distinguish between two subproblems. First, we observe that QRRT and QRRT* have a runtime of $60$s on the Scissor scenario. Both scenarios, however, are ingress scenarios, where the planner needs to find a narrow passage on the base space to enter the goal region, which is challenging for RRT-like algorithms \cite{Kuffner2000} and could be addressed using a bidirectional version of QRRT.

Second, we observe that QMP and QMP* require $14.52$s and $37.27$s to solve the Scissor scenario and that QMP requires $18.98$s to solve the Metal scenario. To explain this rather large increase in runtime, we have a closer look at the individual runtimes, which we show in Table~\ref{table:limitations}. We can observe that both planner exhibit one of two outputs. Either, they quickly return a solution (usually less than $3$s, always less than $10$s) or they fail and time out at $60$s (three/two times for QMP, zero/six times for QMP*). To us, this indicates that both algorithms might be sensitive to the base space path. If the base path is not smooth enough, has kinks in it or is too close to obstacles, then we might not be able to solve it with the pattern dance algorithm. We could address this problem in the future by either additional smoothing of the base space path \cite{Vidal2019}, by introducing conservative heuristics \cite{Chatterjee2019} or by switching to a different relaxed model \cite{Styler2017}.

\subsection{Base path does not admit a feasible section}

While all multilevel planner are probabilistically complete, we often need the pattern dance algorithm to efficiently solve a problem. However, we might encounter scenarios, where the base path does not admit a feasible path section. Such a situation is shown in Fig.~\ref{fig:limitations}. The scenario depicts an X-shape robot, which has to traverse a shape-sorter box with different openings, which we relax by inscribing a sphere (right). Planning for the spherical robot might produce a base path going through the wrong hole. Such a base path does not admit a feasible path section, meaning there are no paths along the path restriction of the base path to traverse towards the goal. While multilevel planner are probabilistically complete and would eventually resolve the situation, we would not be able to solve this situation using our pattern dance algorithm. To address such situations, we could either compute several base paths \cite{Orthey2020WAFR, Ha2019, Vonasek2019, Osa2020, bhattacharya_2018, pokorny_2016_ijrr} and consider them as a multi-arm bandit problem over path restrictions \cite{Kurniawati2008} or we could automatically choose an alternative relaxation using either a meta-heuristic \cite{Brandao2020} or a brute-force search \cite{Orthey2019}.
\section{Conclusion}

We developed the pattern dance algorithm, which takes as input a base space path
and efficiently searches for a feasible section in its path restriction using four dedicated section
patterns, which we named Manhattan, Wriggle, Tunnel and Triple step. We showed in evaluations,
that our pattern dance algorithm successfully coordinates section patterns and outperforms a similar sidestepping algorithm \cite{Orthey2020IJRR}. We then showed that multilevel motion planning algorithms using our pattern dance algorithm outperform classical planner from the OMPL library on challenging narrow passage
scenarios including the Bugtrap, chain egress and precision grasping. With some
exceptions, we often observed runtime improvements by one to two orders of
magnitudes.

While we demonstrated to efficiently solve narrow passage problems, we also
pointed out two limitations. First, we observe an increased runtime in some planning instances. We could address this problem by either optimizing the base path \cite{Zhang2009}, by improved neighborhood modeling \cite{lacevic_2020} or by learning the section patterns themselves \cite{Ichter2018}. Second, we cannot handle cases where the base path does not admit a path section. We could address this problem by
computing multiple base paths \cite{Orthey2020WAFR, Osa2020, Vonasek2019} or using more informed graph restriction sampling methods \cite{Orthey2019}.

Despite limitations, we believe to have contributed a novel solution method
which we can use to efficiently find sections over base path restrictions. We believe our method to be a promising tool to further probe, understand and efficiently exploit high-dimensional state spaces.

\section{Acknowledgement}

Marc Toussaint thanks the   Max Planck Institute for Intelligent Systems for the Max Planck Fellowship.

\begin{IEEEbiography}[{\includegraphics[width=1in,height=1.25in,clip,keepaspectratio]{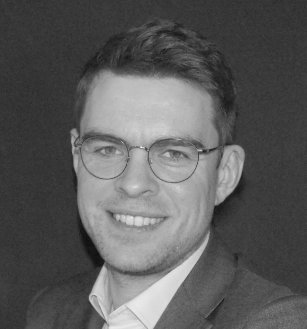}}]{Andreas Orthey}

is a postdoctoral researcher in computational robotics at the TU Berlin funded by the Max Planck Institute for Intelligent Systems (MPI-IS). Previously, he has been a Research Fellow with the Alexander von Humboldt Foundation (AvH) at the University of Stuttgart, the Japan Society for the Promotion of Science (JSPS) at the AIST in Tsukuba, Japan, a Postdoctoral Researcher at the Worcester Polytechnic Institute (WPI), MA, USA and a Doctoral Candidate at the LAAS-CNRS in Toulouse, France. He holds a PhD Degree from INP Toulouse and a Master's Degree with Honours from the TU Berlin. His research interest lies in optimization and planning for complex and high-dimensional robotic systems.
\end{IEEEbiography}
\begin{IEEEbiography}[{\includegraphics[width=1in,height=1.25in,clip,keepaspectratio]{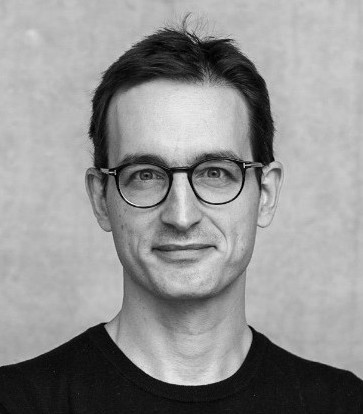}}]{Marc Toussaint}
is professor for Intelligent Systems at TU Berlin since
March 2020 and Max Planck Fellow at the MPI for Intelligent Systems
since November 2018.  In 2017/18 he spend a year as visiting scholar
at MIT, before that some months with Amazon Robotics, and was
professor for Machine Learning and Robotics at the University of
Stuttgart since 2012. In his view, a key in understanding and creating
intelligence is the interplay of learning and reasoning, where
learning becomes the enabler for strongly generalizing reasoning and
acting in our physical world. His research therefore bridges between
AI planning, machine learning, and robotics. His work was awarded best
paper at R:SS'18 and ICMLA'07, and runner up at R:SS'12 and UAI'08.
\end{IEEEbiography}

\bibliographystyle{IEEEtranSN}

{\footnotesize
\balance
\bibliography{IEEEabrv, bib/general, bib/discretesearch}
}

\end{document}